\documentclass[10pt,journal,compsoc]{IEEEtran}
\ifCLASSOPTIONcompsoc
  \usepackage[nocompress]{cite}
\else
  \usepackage{cite}
\fi
\ifCLASSINFOpdf
\else
\fi
\usepackage{amsmath,amssymb,amsfonts}
\usepackage{algorithmic}
\usepackage{graphicx}
\usepackage{textcomp}
\usepackage{url}
\usepackage{color}
\usepackage{caption}
\usepackage{balance}
\usepackage{hyperref}

\begin{document}
\title{Threat of Adversarial Attacks on Deep Learning in Computer Vision: A Survey}

\author{\hspace{30mm} Naveed Akhtar and 
        Ajmal Mian\newline 
        {\footnotesize {\bf ACKNOWLEDGEMENTS:} The authors thank Nicholas Carlini (UC Berkeley) and Dimitris Tsipras (MIT) for feedback to improve the survey quality. We also acknowledge X. Huang (Uni. Liverpool), K. R. Reddy (IISC), E. Valle (UNICAMP), Y. Yoo (CLAIR) and others for providing pointers to make the survey more comprehensive. This research was supported by ARC grant DP160101458.}
        % <-this % stops a space
\IEEEcompsocitemizethanks{\IEEEcompsocthanksitem N. Akhtar and A. Mian are with the School of Computer Science and Software Engineering, University of Western Australia.\protect\\
E-mail: \{naveed.akhtar, ajmal.mian\}@uwa.edu.au
%\IEEEcompsocthanksitem J. Doe and J. Doe are with Anonymous University.
}% <-this % stops an unwanted space
\thanks{Manuscript received August 2017, revised...}}


\markboth{Journal of \LaTeX\ Class Files,~Vol.~PP, August~2017}%
{Shell \MakeLowercase{\textit{et al.}}: Bare Demo of IEEEtran.cls for Computer Society Journals}



\IEEEtitleabstractindextext{%
\begin{abstract}
Deep learning is at the heart of the current rise of artificial intelligence. In the field of Computer Vision, it has become the workhorse for applications ranging from self-driving cars to surveillance and security. Whereas deep neural networks have demonstrated phenomenal success (often beyond human capabilities) in solving complex problems, recent studies show that they are vulnerable to adversarial attacks in the form of  subtle perturbations to inputs that lead a model to predict incorrect outputs.  For  images, such perturbations are often too small to be perceptible, yet they completely fool the deep learning models. Adversarial attacks pose a serious threat to the success of deep learning in practice. This fact has recently lead to a large influx of contributions in this direction. This article presents the first comprehensive survey on adversarial attacks on deep learning in Computer Vision. We  review the works that design adversarial attacks, analyze the existence of such attacks and propose defenses against them. To emphasize that adversarial attacks are possible in practical conditions, we separately review the contributions that evaluate adversarial attacks in the real-world scenarios. Finally, drawing on the reviewed  literature, we provide a broader outlook of this research direction.  
\end{abstract}

}

\maketitle

\IEEEdisplaynontitleabstractindextext

% \fi
\IEEEpeerreviewmaketitle

\IEEEraisesectionheading{\section{Introduction}\label{sec:introduction}}

\IEEEPARstart{D}{eep Learning}~\cite{DL_Nature} is  providing major breakthroughs in solving the problems that have withstood many attempts of machine learning and artificial intelligence community in the past.  As a result, it is currently being used to decipher hard scientific problems at an unprecedented scale, e.g.~in reconstruction of brain circuits~\cite{Mouse}; analysis of mutations in DNA~\cite{Xiong_2015}; prediction of structure-activity of potential drug molecules~\cite{Ma_2015}, and  analyzing the particle accelerator data~\cite{Ciodaro_2012} \cite{Kaggle}. Deep neural networks have also become the preferred choice to solve many challenging tasks in speech recognition~\cite{Hinton_2012} and natural language understanding~\cite{Sutskever_2014}.

In the field of Computer Vision, deep learning became the center of attention after Krizhevsky et al.~\cite{Alex_2012} demonstrated the impressive performance of a Convolutional Neural Network (CNN)~\cite{LeCun1989} based model on a very challenging large-scale visual recognition task~\cite{ImageNet} in 2012.
A significant credit for the current popularity of deep learning can also be attributed to this seminal work.
Since 2012, the Computer Vision community has made numerous valuable contributions to deep learning research, enabling it to provide solutions for the problems encountered in medical science~\cite{MI} to mobile applications~\cite{Mobileapp}. The recent breakthrough in artificial intelligence in the form of tabula-rasa learning of AlphaGo Zero~\cite{AlphaGoZero} also owes a fair share to deep Residual Networks (ResNets)~\cite{ResNet} that were originally proposed for the task of image recognition. 

With the continuous improvements of deep neural network models~\cite{Inceptionv3}, \cite{ResNet}, \cite{Dense}; open access to efficient deep learning software libraries~\cite{MatConvNet}, \cite{Caffe}, \cite{Tensorflow}; and easy availability of hardware required to train complex models, deep learning is fast achieving the maturity to enter into safety and security critical applications, e.g.~self driving cars~\cite{Car}, \cite{Selfdrive2}, surveillance~\cite{Security}, maleware detection~\cite{Papernot_2016a},\cite{Grosse_malware}, drones and robotics~\cite{Voloymer_Human-level},\cite{drone}, and  voice command recognition~\cite{Hinton_2012}.
With the recent real-world developments like facial recognition ATM~\cite{ATM} and Face ID security on mobile phones~\cite{FaceID}, it is apparent that deep learning solutions, especially those originating from Computer Vision problems are about to play a major role in our daily lives.

\begin{figure*}[t] %  figure placement: here, top, bottom, or page
   \centering
   \includegraphics[width=5.5in]{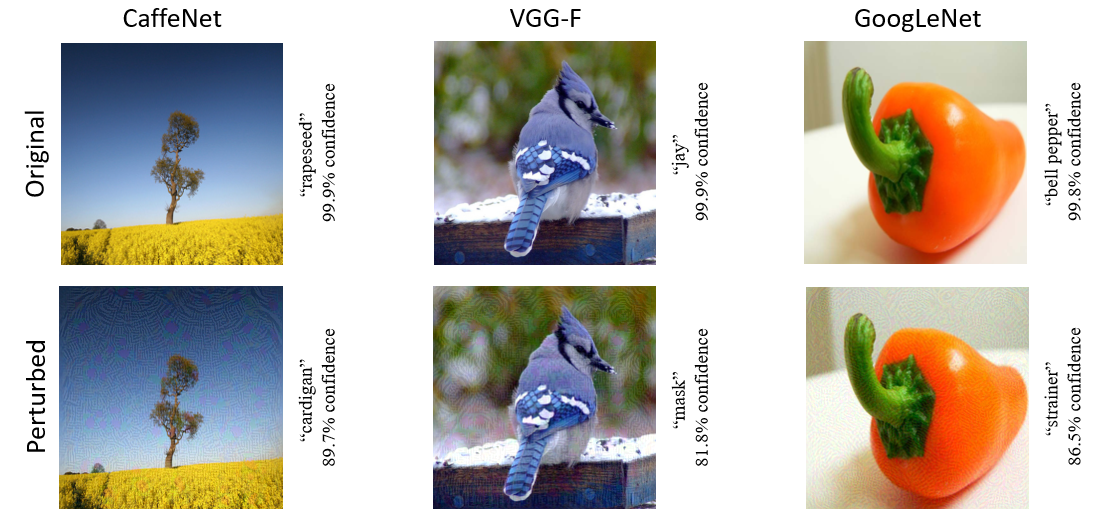} 
   \caption{Example of attacks on deep learning models with `universal adversarial perturbations'~\cite{Uni}: The attacks are shown for the  CaffeNet~\cite{Alex_2012}, VGG-F network~\cite{VGGF} and GoogLeNet~\cite{GoogleNet}. All the networks recognized the original  clean images correctly with high confidence. After small perturbations were added to the images, the networks predicted wrong labels with similar high confidence. Notice that the perturbations are hardly perceptible for human vision system, however their effects on the deep learning models are catastrophic.}  
   \label{fig:teaser}
\end{figure*}

Whereas deep learning performs a wide variety of Computer Vision tasks with remarkable accuracies, Szegedy et al.~\cite{Szegedy_2014} first discovered an intriguing weakness of deep  neural networks in the context of image classification.
They showed that despite their high accuracies, modern deep networks are  surprisingly susceptible to adversarial attacks in the form of small perturbations to images that remain (almost) imperceptible to human vision system.
Such attacks can cause a neural network classifier to completely change its prediction about the image.
Even worse, the attacked models report high confidence on the wrong prediction. Moreover, the same image perturbation can fool multiple network classifiers.
The profound implications of these results triggered a wide interest of researchers in adversarial attacks and their defenses for deep learning in general.

Since the findings of Szegedy et al.~\cite{Szegedy_2014}, several interesting results have surfaced regarding adversarial attacks on deep learning in Computer Vision.
For instance,  in addition to the image-specific adversarial perturbations~\cite{Szegedy_2014}, Moosavi-Dezfooli et al.~\cite{Uni} showed the existence of `universal perturbations' that can fool a network classifier on {\it any} image (see Fig.~\ref{fig:teaser} for example). Similarly, Athalye et al.~\cite{Athalye_2017} demonstrated that it is possible to even 3-D print real-world objects that can fool deep neural network classifiers (see Section~\ref{sec:3d}).
Keeping in view the significance of deep learning research in Computer Vision and its potential applications in the real life, this article presents the first comprehensive survey on adversarial attacks on deep learning in Computer Vision.
The article is intended for a wider readership than Computer Vision community, hence it assumes only basic knowledge of deep learning and image processing. Nevertheless, it also discusses technical details of important contributions for the interested readers.

We first describe the common terms related to adversarial attacks in the parlance of Computer Vision in Section~\ref{sec:Def}. In Section~\ref{sec:Attacks}, we review the adversarial attacks for the task of image classification and beyond.
A separate section is dedicated to the approaches that deal with adversarial attacks in the real-world conditions. Those approaches are reviewed in Section~\ref{sec:Real}.
In the literature, there are also works that mainly focus on analyzing the existence of adversarial attacks. We discuss those contributions in Section~\ref{sec:exist}.
The approaches that make defense against the adversarial attacks as their central topic are discussed in Section~\ref{sec:Defense}. In Section~\ref{sec:future}, we provide a broader outlook of the research direction based on the reviewed literature. Finally, we draw conclusion in Section~\ref{sec:conc}.

%- See Carlini and wagner paper for introduction writing... there are attacks in voice recog and other places.

%Towards the Science of Security and Privacy in Machine Learning \cite{Papernot_2016a}.

\section{Definitions of terms}
\label{sec:Def}
In this section, we describe the common technical terms  used in the literature related to adversarial attacks on deep learning in Computer Vision.
The remaining article also follows the same definitions of the terms.  
\begin{itemize}
\item {\it Adversarial example/image} is a modified version of a clean image that is intentionally perturbed (e.g.~by adding noise) to confuse/fool a machine learning technique, such as deep neural networks.
\item {\it Adversarial perturbation} is the noise that is added to the clean image to make it an adversarial example.
\item {\it Adversarial training} uses adversarial images besides the clean images to train machine learning models.  
\item {\it Adversary} more commonly refers to the agent who creates an adversarial example. However, in some cases the example itself is also called adversary.
\item {\it Black-box attacks} feed a targeted model with the adversarial examples (during testing) that are generated without the knowledge of that model. In some instances, it is assumed that the adversary has a limited knowledge of the model (e.g.~its training procedure and/or its architecture) but definitely does not know about the model parameters. In other instances, using any information about the target model is referred to as `semi-black-box' attack. We use the former convention in this article.
\item {\it Detector} is a mechanism to (only) detect if an image is an adversarial example.
\item {\it Fooling ratio/rate} indicates the percentage of images on which a trained model changes its prediction label after the images are perturbed.
\item {\it One-shot/one-step methods} generate an adversarial perturbation by performing a single step computation, e.g. computing gradient of model loss once. The opposite are iterative methods that perform the same computation multiple times to get a single perturbation. The latter are often  computationally expensive.
\item {\it Quasi-imperceptible} perturbations impair images very slightly for human perception. 
\item {\it Rectifier} modifies an adversarial example to restore the prediction of the targeted model to its prediction on the clean version of the same example.
\item {\it Targeted attacks} fool a model into falsely predicting a specific label for the adversarial image. They are opposite to the {\it non-targeted} attacks in which the predicted label of the adversarial image is irrelevant, as long as it is not the correct label.  
\item {\it Threat model} refers to the types of potential attacks considered by an approach, e.g. black-box attack.
\item {\it Transferability} refers to the ability of an adversarial example to remain effective even for the models other than the one used to generate it.  
\item {\it Universal perturbation} is able to fool a given model on `any' image with high probability.  Note that, universality refers to the property of a  perturbation being `image-agnostic' as opposed to having good transferability.
\item {\it White-box attacks} assume the complete knowledge of the targeted model, including its parameter values, architecture, training method, and in some cases its training data as well.
\end{itemize}

% - Adversary (both people and examples in some context)
% - Adversarial perturbation
% - Adversarial example
% - Adversarial training
% - blackbox attacks
% - Detector
% - Fooling ratio/rate
% -(may be) One shot or one step methods 
% - Quasi-imperceptible
% - Rectifier
% - Targeted attacks
% - Threat model
% - transferability
% - Universal PErturbations - doubly-universal
% - Whitebox attacks

% \section{Interesting Observations/claims/Guidelines}
% Keep an eye on anything particularly important in abstracts etc.

% - Goodfellow et al.~\cite{Goodfellow_2015} say its too linear. Tany and Griffin \cite{Tanay_2016} construct linear classifiers that can not be fooled. Again \cite{Papernot_2016c} say that the linearity may be true but \cite{Su_2017} argue against it.

% -In univesal perturbations there is a case of dominant labels

\section{Adversarial attacks}
\label{sec:Attacks}
In this section, we review the body of literature in Computer Vision that introduces methods for adversarial attacks on deep learning.
The reviewed literature mainly deals with the art of fooling the deep neural networks in `laboratory settings', where  approaches are developed for the typical Computer Vision tasks, e.g.~recognition, and their effectiveness is demonstrated using  standard datasets, e.g.~MNIST~\cite{LeCun1989}.
The techniques that  focus on attacking deep learning in the real-world conditions are separately reviewed in Section~\ref{sec:Real}.
However, it should be noted that the approaches reviewed in this section form the basis of the real-world attacks, and almost each one of them has the potential to significantly affect deep learning in practice.
Our division is  based on the evaluation conditions of the attacks in the original contributions.

The review in this section is mainly organized in chronological order, with few exceptions to maintain the flow of discussion. To  provide technical understanding of the core concepts to the reader, we also go into technical details of  the popular approaches and some representative techniques of the emerging directions in this area. Other methods are discussed briefly. We refer to the original papers for the details on those techniques. This section is divided into two parts.
In part~\ref{sec:Popular}, we review the methods that attack deep neural networks performing the most common task in Computer Vision, i.e.~classification/recognition. Approaches that are predominantly designed to attack deep learning beyond this task are discussed in part~\ref{sec:beyondClassification}.

\subsection{Attacks for classification}
\label{sec:Popular}
%\noindent{\bf Box-constrained L-BFGS:}
\subsubsection{Box-constrained L-BFGS} Szegedy et al.~\cite{Szegedy_2014} first demonstrated the existence of small perturbations to the images, such that the perturbed images could fool  deep learning models into misclassification.
Let ${\bf I}_c \in \mathbb R^m$ denote a vectorized clean image - the subscript `$c$' emphasizes that the image is clean. To compute an additive  perturbation $\boldsymbol\rho \in \mathbb R^m$ that would distort the image very slightly to fool the network, Szegedy et al.~proposed to solve the following problem: 
\begin{align}
\min_{\boldsymbol\rho} || \boldsymbol\rho ||_2 \hspace{2mm} \text{s.t.}~\mathcal C( {\bf I}_c  +  \boldsymbol\rho) = \ell;~{\bf I}_c  +  \boldsymbol\rho \in [0,1]^m,  
\label{eq:Intriguing}
\end{align}
where `$\ell$' denotes the label of the image and $\mathcal C(.)$ is the deep neural network  classifier. The authors proposed to solve (\ref{eq:Intriguing}) for its non-trivial solution where `$\ell$' is different from the original label of ${\bf I}_c$. In that case, (\ref{eq:Intriguing}) becomes a hard problem, hence an approximate solution is sought using a box-constrained L-BFGS~\cite{BFGS}. This is done by finding the minimum $c>0$ for which the minimizer $\boldsymbol\rho$ of the following problem satisfies the condition $\mathcal C({\bf I}_c  +  \boldsymbol\rho) = \ell$:
\begin{align}
\min_{\boldsymbol\rho}~~c|\boldsymbol\rho| + \mathcal{L}({\bf I}_c  +  \boldsymbol\rho, \ell)~~s.t.~{\bf I}_c  +  \boldsymbol\rho \in [0,1]^m,
\label{eq:approxSzegedy}
\end{align}
where $\mathcal L(.,.)$ computes the loss of the classifier. We note that (\ref{eq:approxSzegedy}) results in the exact solution for a classifier that has a convex loss function. However, for deep neural networks, this is generally not the case.  The computed perturbation is simply added to the image to make it an adversarial example.

\begin{figure}[t] %  figure placement: here, top, bottom, or page
   \centering
   \includegraphics[width=2.5in]{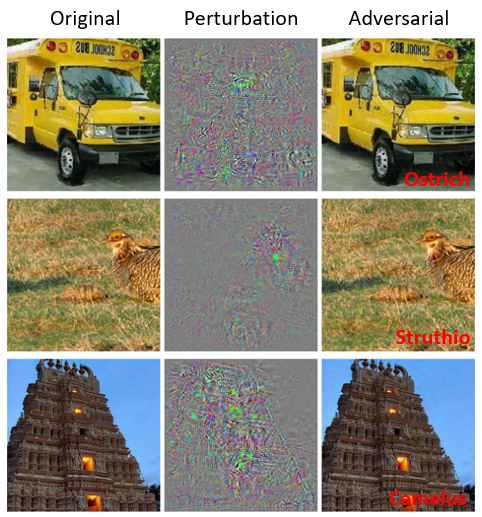} 
   \caption{Illustration of adversarial examples generated using~\cite{Szegedy_2014} for AlexNet~\cite{Alex_2012}. The perturbations are magnified 10x for better visualization (values shifted by 128 and clamped). The predicted labels of adversarial examples are also shown.}  
   \label{fig:IntriguingExp}
\end{figure}

As shown in Fig.~\ref{fig:IntriguingExp}, the above method is able to compute perturbations that when added to clean images fool a neural network, but the adversarial images appear similar to the clean images to the human vision system.
It was observed by Szegedy et al.~that the perturbations computed for one neural network were also able to fool multiple networks. These astonishing results identified a blind-spot in  deep learning. 
At the time of this discovery the Computer Vision community was fast adapting to the impression that deep learning features define the space where perceptual distances are well approximated by the Euclidean distances. Hence, these contradictory results triggered a wide interest of researchers in adversarial attacks on deep learning in Computer Vision.

\subsubsection{Fast Gradient Sign Method (FGSM)}
%\noindent{\bf Fast Gradient Sign Method (FGSM)}\cite{Goodfellow_2015}: 
It was observed by Szegedy et al.~\cite{Szegedy_2014} that the robustness of deep neural networks against the adversarial examples could be improved by adversarial training. To enable effective adversarial training, Goodfellow et al.~\cite{Goodfellow_2015} developed a method to efficiently compute an adversarial perturbation for a given image by solving the following problem:
\begin{align}
\boldsymbol\rho = \epsilon~\text{sign}\left( \nabla \mathcal J (\boldsymbol\theta, {\bf I}_c, \ell)  \right),
\label{eq:FGSM}
\end{align}
where $\nabla \mathcal J(.,.,.)$ computes the gradient of the cost function around the current value of the model parameters $\boldsymbol\theta$ w.r.t.~${\bf I}_c$, sign$(.)$ denotes the sign function and $\epsilon$ is a small scalar value that restricts the  norm of the perturbation. The method for solving (\ref{eq:FGSM}) was termed `Fast Gradient Sign Method' (FGSM) in the original work.

Interestingly, the adversarial examples generated by FGSM exploit the `linearity' of deep network models in the higher dimensional space whereas such models were commonly thought to be highly non-linear at that time. 
Goodfellow et al.~\cite{Goodfellow_2015} hypothesized that the designs of modern deep neural networks that (intentionally) encourage linear behavior for computational gains, also make them susceptible to cheap analytical perturbations. In the related  literature, this idea is often referred to as the `linearity hypothesis', which is substantiated by the FGSM approach. 

Kurakin et al.~\cite{Kurakin_scale} noted that on the popular large-scale image recognition data set ImageNet~\cite{ImageNet}, the top-1 error rate on the adversarial examples generated by FGSM is around $63 -69\%$ for $\epsilon \in [2, 32]$. The authors also proposed a `one-step target class' variation of the FGSM where instead of using the true label $\ell$ of the image in (\ref{eq:FGSM}), they used the label $\ell_{\text{target}}$ of the least likely class predicted by the network for ${\bf I}_c$. The computed perturbation is then subtracted from the original image to make it an adversarial example. For a neural network with cross-entropy loss, doing so maximizes the probability that the network predicts $\ell_{\text{target}}$ as the label of the adversarial example. 
It is suggested, that a random class can also be used as the target class for fooling the network, however it may lead to less interesting fooling, e.g.~misclassification of one breed of dog as another dog breed. 
The authors also demonstrated that adversarial training improves robustness of deep neural networks against the attacks generated by  FGSM and its proposed variants.  

%FGSM essentially increases the loss of the classifier for the given image 
The FGSM  perturbs an image to increase the loss of the classifier on the resulting image. The sign function ensures that the magnitude of the loss is maximized, while $\epsilon$ essentially restricts the $\ell_{\infty}$-norm of the perturbation.
%and uses the signum function and $\epsilon$ to maximize this loss such that the $\ell_{\infty}$-norm of the perturbation is clipped at $\epsilon$. 
Miyato et al.~\cite{VAT} proposed a closely related method to compute the perturbation as follows
\begin{align}\boldsymbol\rho = \epsilon\frac{\nabla \mathcal J (\boldsymbol\theta, {\bf I}_c, \ell)}{ || \nabla \mathcal J (\boldsymbol\theta, {\bf I}_c, \ell) ||_2}.
\end{align}
In the above equation, the computed gradient is normalized with its $\ell_2$-norm. Kurakin et al.~\cite{Kurakin_scale} referred to this technique as `Fast Gradient L$_2$' method and also proposed an alternative of  using the $\ell_{\infty}$-norm for normalization, and referred to the resulting technique as `Fast Gradient L$_\infty$' method. Broadly speaking, all of these methods are seen as `one-step' or `one-shot' methods in the literature related to adversarial attacks in Computer Vision.

\subsubsection{Basic \& Least-Likely-Class Iterative Methods}
%\noindent{\bf Basic \& Least-Likely-Class Iterative Methods}~\cite{Kurakin_2016a}:
The one-step methods perturb images by taking a single large step in the direction that increases the loss of the classifier (i.e.~one-step gradient ascent). An intuitive extension of this idea is to iteratively take multiple small steps while adjusting the direction after each step. The Basic Iterative Method (BIM)~\cite{Kurakin_2016a} does exactly that, and iteratively computes the following:
\begin{align}
{\bf I}_{\boldsymbol\rho}^{i + 1} = \text{Clip}_{\epsilon} \left \{ {\bf I}_{\boldsymbol\rho}^i + \alpha~\text{sign} (\nabla \mathcal J(\boldsymbol\theta, {\bf I}_{\boldsymbol\rho}^i, \ell) \right \},
\label{eq:BIM}
\end{align}
where ${\bf I}_{\boldsymbol\rho}^i$ denotes the perturbed image at the $i^{\text{th}}$ iteration, Clip$_\epsilon \{.\}$ clips (the values of the pixels of) the image in its argument at $\epsilon$ and $\alpha$ determines the step size (normally, $\alpha =1$). The BIM algorithm starts with ${\bf I}_{\boldsymbol\rho}^0 = {\bf I}_{c}$ and runs for the number of iterations determined by the formula $\lfloor \min (\epsilon +4, 1.25 \epsilon)\rfloor$. 
Madry et al.~\cite{Madry_2017} pointed out that BIM is equivalent to (the $\ell_\infty$ version of) Projected Gradient Descent (PGD), a standard convex optimization method. 

Similar to extending the FGSM to its `one-step target class' variation, Kurakin et al.~\cite{Kurakin_2016a} also extended BIM to Iterative Least-likely Class Method (ILCM). In that case, the label $\ell$ of the image in (\ref{eq:BIM}) is replaced by the target label $\ell_{\text{target}}$ of the least likely class predicted by the classifier. 
The adversarial examples generated by the ILCM method has been shown to seriously affect the classification accuracy of a modern deep architecture  Inception v3~\cite{Inceptionv3}, even for very small values of $\epsilon$, e.g. $< 16$.

\subsubsection{Jacobian-based Saliency Map Attack (JSMA)}
%\noindent{\bf Jacobian-based Saliency Map Attack (JSMA)}~\cite{Papernot_2016c}:
In the literature, it is more common to generate adversarial examples by restricting   $\ell_{\infty}$ or $\ell_2$-norms of the perturbations to make them imperceptible for humans. However, Papernot et al.~\cite{Papernot_2016c} also created an adversarial attack by restricting the $\ell_0$-norm of the perturbations.
Physically, it means that the goal  is to modify only a few pixels in the image instead of perturbing the whole image to fool the classifier.
The crux of their algorithm to generate the desired adversarial image can be understood as follows.
The algorithm modifies pixels of the clean image one at a time and monitors the effects of the change on the resulting classification. The monitoring is performed by computing a saliency map using the gradients of the outputs of the network layers. In this map, a larger value indicates a higher likelihood of fooling the network to predict $\ell_{\text{target}}$ as the label of the modified image instead of the original label $\ell$.
Thus, the algorithm performs targeted fooling. 
Once the map has been computed, the algorithm chooses the pixel that is most effective to fool the network and alters it. This process is repeated until either the maximum number of allowable pixels are altered in the adversarial image or the fooling succeeds.

%Given the computed saliency map and the image, the algorithm iteratively chooses the current most effective pixel for network fooling and modifies it to form adversarial image.

\subsubsection{ One Pixel Attack}
\begin{figure}[t] %  figure placement: here, top, bottom, or page
   \centering
   \includegraphics[width=3.3in]{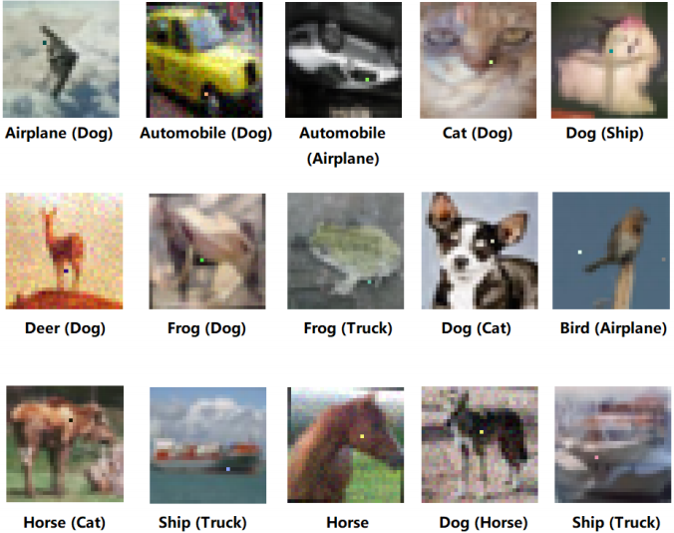} 
   \caption{Illustration of one pixel adversarial attacks~\cite{Su_2017}: The correct label is mentioned with each image. The corresponding predicted label is given in parentheses. }  
   \label{fig:OnePixel}
\end{figure}

%\noindent{\bf One Pixel Attack}~\cite{Su_2017}: 
An extreme case for the adversarial attack is when only one pixel in the image is changed to fool the classifier. Interestingly, Su et al.~\cite{Su_2017} claimed successful fooling of three different network models on $70.97\%$ of the tested images by changing just one pixel per image. They also reported that the average confidence of the networks on the wrong labels was  found to be  $97.47\%$. 
We show representative examples of the adversarial images from \cite{Su_2017} in Fig.~\ref{fig:OnePixel}.
Su et al. computed the adversarial examples by using the concept  of  Differential Evolution~\cite{DiffEval}.
For a clean image ${\bf I}_c$, they first created a set of $400$ vectors in $\mathbb R^5$ such that each vector contained $xy$-coordinates and RGB values for an arbitrary candidate pixel. Then, they randomly modified the elements of the vectors to create children such that a child competes with its parent for fitness in the next iteration, while the probabilistic predicted label of the network is used as the fitness criterion.  The last surviving child is used to alter the pixel in the image.

Even with such a simple evolutionary strategy Su et al.~\cite{Su_2017} were able to show successful fooling of deep networks. 
Notice that,  differential evolution enables their approach to generate adversarial examples without having access to any information about the network parameter values or their gradients. The only input their technique requires is the probabilistic labels predicted  by  the targeted model.
%Therefore, they termed the attacks created by their approach as semi-black box attacks.

\subsubsection{Carlini and Wagner Attacks (C\&W)}
%\noindent{\bf Carlini and Wagner Attacks (C\&W)}~\cite{Carlini_2016}: 
A set of three adversarial attacks were introduced by Carlini and Wagner~\cite{Carlini_2016} in the wake of defensive distillation against the adversarial perturbations \cite{Papernot_2016}. These attacks make the perturbations quasi-imperceptible by restricting their $\ell_2$, $\ell_{\infty}$ and $\ell_0$ norms, and it is shown that defensive distillation for the targeted networks almost completely fails against these attacks. Moreover, it is also shown that the adversarial examples generated using the unsecured (un-distilled) networks transfer well to the secured (distilled) networks, which makes the computed perturbations suitable for black-box attacks.

Whereas it is more common to exploit the transferability property of adversarial examples to generate black-box attacks,  Chen et al.~\cite{ZOO} also proposed `Zeroth Order Optimization (ZOO)' based attacks that directly estimate the gradients of the targeted model for generating the adversarial examples. 
These attacks were  inspired by C\&W attacks. We refer to the original papers for further details on C\&W and ZOO attacks.

\subsubsection{ DeepFool}
%\noindent{\bf DeepFool}~\cite{DeepFool}: 
Moosavi-Dezfooli et al.~\cite{DeepFool} proposed to  compute a minimal norm adversarial perturbation for a given image in an iterative manner.
Their algorithm, i.e.~DeepFool initializes with the clean image that is assumed to reside in a region confined by the decision boundaries of the classifier. This region decides the class-label of the image. At each iteration, the algorithm perturbs the  image by a small vector that is computed to take the resulting image to the boundary of the polyhydron that is obtained by linearizing the boundaries of the region within which the image resides. The  perturbations added to the image in each iteration are accumulated to compute the final perturbation once the perturbed image changes its label according to the original decision boundaries of the network.
The authors show that the DeepFool algorithm is able to compute perturbations that are smaller than the perturbations computed by FGSM~\cite{Goodfellow_2015} in terms of their norm, while having similar fooling ratios. 

\subsubsection{Universal Adversarial Perturbations}
%\noindent{\bf Universal Adversarial Perturbations}~\cite{Uni}: 
Whereas the methods like FGSM~\cite{Goodfellow_2015}, ILCM~\cite{Kurakin_2016a}, DeepFool~\cite{DeepFool} etc.~compute perturbations to fool a network on a single image, the `universal' adversarial perturbations computed by Moosavi-Dezfooli et al.~\cite{Uni} are able to fool a network on `any' image with high probability. These image-agnostic perturbations also remain quasi-imperceptible for the human vision system, as can be observed in Fig.~\ref{fig:teaser}. To formally define these  perturbations, let us assume that clean images are sampled from the distribution $\boldsymbol\Im_c$. A perturbation $\boldsymbol\rho$ is `universal' if it satisfies the following constraint:
\begin{align}
\underset{{\bf I}_c \sim \boldsymbol\Im_c}{\mathrm{\text P}} \Big( \mathcal C({\bf I}_c) \neq \mathcal C({\bf I}_c + \boldsymbol \rho) \Big) \geq \delta ~~~\text{s.t.}~~||\boldsymbol\rho||_p \leq \xi,
\label{eq:def1}
\end{align}
where P(.) denotes the probability, $\delta \in (0, 1]$ is the fooling ratio, $||.||_p$ denotes the $\ell_p$-norm and $\xi$ is a pre-defined constant. The smaller the value of $\xi$, the harder it is to perceive the perturbation in the image. Strictly speaking, the perturbations that satisfy (\ref{eq:def1}) should be referred to as $(\delta, \xi)$-universal because of their  strong dependence on the mentioned parameters. However, these perturbations are commonly referred to as the `universal adversarial perturbations' in the literature. 

The authors computed the universal perturbations by restricting their $\ell_2$-norm as well as $\ell_{\infty}$-norm, and showed that the perturbations with their norms upper bounded by $4\%$ of the respective image norms already achieved significant fooling ratios of around 0.8 or more for state-of-the-art image classifiers. Their iterative approach to compute a perturbation is related to the DeepFool strategy~\cite{DeepFool} of gradually pushing a data point (i.e.~an image) to the decision boundary for its class. However, in this case, `all' the training data points are sequentially pushed to the respective decision boundaries and the perturbations computed over all the images are gradually accumulated by back-projecting the accumulator to the desired $\ell_p$ ball of radius $\xi$ every time.

The algorithm proposed by Moosavi-Dezfooli et al.~\cite{Uni} computes perturbations while targeting a single network model, e.g. ResNet~\cite{ResNet}. However, it is shown that these perturbations also generalize well across different networks (especially those having similar architectures). In that sense, the author's claim the perturbations to be, to some extent, `doubly universal'.
Moreover, it is also shown that high fooling ratio (e.g.~$\delta \geq  0.5$) is achievable by learning a perturbation using only around $2,000$ training images.

Khrulkov et al.~\cite{SingularVector} also proposed a method for constructing universal adversarial perturbations as singular vectors of the Jacobian matrices of feature maps of the networks, which allowed for achieving relatively high fooling rates using only a small number of images. Another method to generate universal perturbations is fast-feature-fool by Mopuri et al.~\cite{Mopuri_2017fff}. Their method generates the universal perturbations independent of data.

\subsubsection{UPSET and ANGRI}
Sarkar et al.~\cite{Sarkar_2017UPSET} proposed two black-box attack algorithms, namely  UPSET: Universal Perturbations for Steering to Exact Targets, and
ANGRI: Antagonistic Network for Generating Rogue Images for targeted fooling of deep  neural networks.
For `n' classes, UPSET seeks to produce `n' image-agnostic perturbations such
that when the perturbation is added to an image that does not belong to a targeted class, the classifier will classify the perturbed image as
being from that class. The power of UPSET comes from a residual generating network R(.), that takes the target class `t' as input and produces a perturbation R(t) for fooling. The overall method solves the following optimization problem using the so-called UPSET network:
\begin{align}
{\bf I}_{\boldsymbol\rho} = \max(\min(s \text{R(t)} + {\bf I}_c, 1), -1),
\end{align}
where the pixel values in ${\bf I}_c$ are normalized to lie in $[-1, 1]$, and `$s$' is a scalar. To ensure ${\bf I}_{\boldsymbol\rho}$ to be a valid image, all values outside the interval $[-1, 1]$ are clipped.
As compared to the image-agnostic perturbations of UPSET, ANGRI computes image-specific perturbations in a closely  related manner, for which we refer to the original work. The perturbations resulting from ANGRI are also used for targeted fooling. 
Both algorithms have been reported to achieve high fooling ratios on MNIST~\cite{LeCun1989} and CIFAR-10~\cite{CIFAR} datasets.

\subsubsection{Houdini}
%\noindent {\bf Houdini}~\cite{Cisse_2017}: 
Cisse et al.~\cite{Cisse_2017} proposed `Houdini'- an approach for fooling gradient-based learning machines by generating adversarial examples that can be tailored to task losses.
Typical algorithms that generate adversarial examples employ gradients of differentiable loss functions of the networks to compute the perturbations. However, task losses are often not amenable to this approach. For instance, the task loss of speech recognition is based on word-error-rate, which does not allow straightforward exploitation of loss function gradient. Houdini is tailored to generate adversarial examples for such tasks. Besides successful generation of adversarial images for classification, Houdini has also  been shown to  successfully attack a popular deep Automatic Speech Recognition system~\cite{DeepSpeech2}.
The authors have also demonstrated the transferability of attacks in speech recognition by fooling Google Voice in a black-box attack scenario.
Moreover, successful targeted and non-targeted attacks are also demonstrated for a deep learning model for human pose estimation.

\subsubsection{Adversarial Transformation Networks (ATNs)}
Baluja and Fischer~\cite{Baluja_2017}   trained feed-forward neural networks to generate adversarial examples against other targeted networks or set of networks. The trained models were termed Adversarial Transformation Networks (ATNs). The adversarial examples generated by these networks are computed by minimizing a joint loss function comprising of two parts. The first part restricts the adversarial example to have perceptual similarity with the original image, whereas the second part aims at altering the prediction of the targeted network on the resulting image.

Along the same direction, Hayex and Danezis~\cite{Service} also used an attacker neural network to learn adversarial examples for black-box attacks. In the presented results, the examples computed by the attacker network remain perceptually indistinguishable from the clean images but they are misclassified by the targeted networks with overwhelming probabilities - reducing classification accuracy from 99.4\% to 0.77\% on MNIST data~\cite{LeCun1989}, and from 91.4\% to 6.8\% on the CIFAR-10 dataset~\cite{CIFAR}.

\begin{table*}[t]
\centering
\caption{Summary of the  attributes of diverse attacking methods: The `perturbation norm' indicates the restricted $\ell_p$-norm of the perturbations to make them imperceptible. The strength (higher for more asterisks) is based on the impression from the reviewed literature.   }
\label{tab:1}
\begin{tabular}{|l||c|c|c|c|c|c|}
\hline
 {\bf Method}							&  Black/White box	&  Targeted/Non-targeted 	& Specific/Universal 	& Perturbation norm & Learning 		& Strength  \\ \hline\hline
L-BFGS~\cite{Szegedy_2014} 		&  White box		&  Targeted					& Image specific			& $\ell_{\infty}$ 	& One shot		& $***$	\\ \hline
FGSM~\cite{Goodfellow_2015}		&  White box		&  Targeted					& Image specific			& $\ell_{\infty}$	& One shot		& $***$ \\ \hline
BIM \& ILCM~\cite{Kurakin_2016a}&  White box		& Non targeted				& Image specific			& $\ell_{\infty}$	& Iterative		& $**$$**$ 	\\ \hline
JSMA~\cite{Papernot_2016c}		&  White box		& Targeted					& Image specific			& $\ell_{0}$	& Iterative			& $***$ 	\\ \hline
One-pixel~\cite{Su_2017}	& Black box		& Non Targeted		& Image specific		& $\ell_0$		& Iterative	& $**$ \\ \hline
C\&W attacks~\cite{Carlini_2016} & White box	& Targeted	&	Image specific		&	$\ell_0, \ell_2, \ell_{\infty}$	&	Iterative & $*****$ \\ \hline
DeepFool~\cite{DeepFool} & White box	& Non targeted		& Image specific		& $\ell_2, \ell_{\infty}$	& Iterative	& $**$$**$ \\ \hline
Uni. perturbations~\cite{Uni} & White box	& Non targeted & Universal &	$\ell_2, \ell_{\infty}$ & Iterative & $*****$ \\ \hline
UPSET~\cite{Sarkar_2017UPSET} & Black box	& Targeted		& Universal 	&  $\ell_{\infty}$ & Iterative & $**$$**$ \\ \hline
ANGRI~\cite{Sarkar_2017UPSET} & Black box	& Targeted		& Image specific	& $\ell_{\infty}$ &	Iterative & $**$$**$ \\ \hline
Houdini~\cite{Cisse_2017} & Black box		& Targeted		& Image specific	& $\ell_2, \ell_{\infty}$ & Iterative & $**$$**$ \\ \hline
ATNs~\cite{Baluja_2017} & White box			& Targeted		& Image specific	& $\ell_{\infty}$ & Iterative & $**$$**$ \\ \hline
\end{tabular}
\end{table*}

\subsubsection{Miscellaneous Attacks} 
The adversarial attacks discussed above are either the popular ones in the recent literature or they are representative of the research directions that are fast becoming popular. 
A summary of the main attributes of these attacks is also provided in Table~\ref{tab:1}.
For a comprehensive study, below we provide brief descriptions of further techniques to generate adversarial attacks on deep neural networks. We note that this research area is currently highly active. Whereas every attempt has been made to review as many approaches as possible, we do not claim the review to be exhaustive. Due to high activity in this research direction, many more attacks are likely to surface in the near future. 

Sabour et al.~\cite{Sabour_2015} showed the possibility of generating adversarial examples by altering the internal layers of deep neural networks. The authors demonstrated that it is  possible to make internal network representation of adversarial images to resemble  representations of images from different classes. Papernot et al.~\cite{Papernot_transfer} studied transferability of adversarial attacks for deep learning as well as other machine learning techniques and introduced further transferability attacks. Narodytska and Kasiviswanathan~\cite{Narodytska_2016} also introduced further black-box attacks that have been found effective in fooling the neural networks by changing only few pixel values in the images.
Liu et al.~\cite{Liu_2017} introduced `epsilon-neighborhood' attack that have been shown to fool defensively distilled networks~\cite{Papernot_2016On} with $100\%$ success for white-box attacks.
Oh et al.~\cite{Oh_2017} took a `Game Theory' perspective on adversarial attacks and derived a strategy to counter the counter-measures taken against adversarial attacks on deep neural networks.
Mpouri et al.~\cite{Mopuri_2017fff} developed a data-independent approach to generate universal adversarial perturbations for the deep network models.
Hosseini et al.~\cite{Hosseini_2017b} introduced the notion of `semantic adversarial examples' - input images that represent semantically same objects for humans but deep neural networks misclassify them. They used negatives of the images as semantic adversarial examples. Kanbak et al.~\cite{Kanbak_2017} introduced `ManiFool' algorithm in the wake of DeepFool method~\cite{DeepFool} to measure robustness of deep neural networks against geometrically perturbed images.
Dong et al.~\cite{Dong_2017} proposed an iterative method to boost adversarial attacks for black-box scenarios.
Recently, Carlini and Wagner~\cite{Carlini_2017b} also demonstrated that ten different defenses against perturbations can again be defeated by new attacks constructed using new loss functions.
Rozsa et al.~\cite{Rozsa_2016} also proposed a `hot/cold' method for generating multiple possible adversarial examples for a single image. 
Interestingly, adversarial perturbations are not only being added to images to reduces the accuracy of deep learning classifiers. Yoo et al.~\cite{Butterfly} recently proposed an approach to also slightly improve the classification performance with the help of subtle perturbation to images. 

We note that the authors of many works reviewed in this article have made the source code of their implementations publicly available. This is one of the major reasons behind the current rise in this research direction. Beside those resources, there are also libraries, e.g. Cleverhans~\cite{Papernot_CleverHans1}, \cite{Goodfellow_CleverHans0} that have started emerging  in order to further boost this research direction.
Adversarial-Playground (\url{https:
//github.com/QData/AdversarialDNN-Playground}) is another example of a toolbox made public by Norton and Qi~\cite{Norton_2017playground} to understand adversarial attacks. 

\subsection{Attacks beyond classification/recognition}
\label{sec:beyondClassification}
With the exception of Houdini~\cite{Cisse_2017}, all the mainstream adversarial attacks reviewed in Section~\ref{sec:Popular} directly focused on the task of classification - typically fooling CNN-based~\cite{LeCun1989} classifiers. However, due to the seriousness of adversarial threats, attacks are also being actively investigated beyond the  classification/recognition task in Computer Vision. Below, we review the works that develop approaches to attack deep neural networks beyond classification.  

%\noindent{\bf Attacks on Autoencoders and Generative Models}:  
\subsubsection{Attacks on Autoencoders and Generative Models}  
Tabacof et al.~\cite{Tabacof_2016a} investigated adversarial attacks for autoencoders~\cite{AE}, and proposed a technique to distort input image (to make it adversarial) that misleads the autoencoder to reconstruct a completely different image. 
Their approach attacks the internal representation of a neural network such that the representation for the adversarial image becomes similar to that of the target image.
However, it is reported in~\cite{Tabacof_2016a} that autoencoders seem to be  much more robust to adversarial attacks than the typical classifier networks.
Kos et al.~\cite{Kos_2017} also explored methods for computing adversarial examples for deep generative models, e.g.~variational autoencoder (VAE) and the VAE-Generative Adversarial Networks (VAE-GANs). GANs, such as \cite{GANs} are becoming exceedingly popular now-a-days in Computer Vision applications due to their ability to learn data distributions and generate realistic images using those distributions. The authors introduced three different classes of attacks for VAE and VAE-GANs.  
Owing to the success of these attacks it is concluded that the deep generative models are also vulnerable to adversaries that can convince them to turn inputs into very different outputs.
This work adds further support to the hypothesis that ``adversarial examples are a general phenomenon for current neural network architectures''. 

\subsubsection{Attack on Recurrent Neural Networks} 
%\noindent{\bf Attack on Recurrent Neural Networks}:  
Papernot et al.~\cite{Papernot_craft} successfully generated adversarial  input sequences for Recurrent Neural Networks (RNNs). RNNs are deep learning models that are particularly suitable for learning mappings between sequential inputs and outputs~\cite{RNN}. Papernot et al. demonstrated that the algorithms proposed to compute adversarial examples for the feed-forward neural networks (e.g.~FGSM~\cite{Goodfellow_2015}) can also be adapted for fooling RNNs. In particular, the authors demonstrated successful fooling of the popular Long-Short-Term-Memory (LSTM) RNN architecture~\cite{LSTM}. It is concluded that the cyclic neural network model like RNNs are also not immune to the adversarial perturbations that were originally uncovered in the context of acyclic neural networks, i.e.~CNNs.

\subsubsection{Attacks on Deep Reinforcement Learning} 
Lin et al.~\cite{Lin_2017Tactics} proposed two different adversarial attacks for the agents trained by deep reinforcement learning~\cite{Voloymer_Human-level}. In the first attack, called `strategically-timed attack', the adversary
minimizes the reward of the agent by attacking it at a small subset of time steps in an episode. 
A method is proposed to determine when an adversarial example should
be crafted and applied, which enables the attack to go undetected. In the second attack, referred as `enchanting attack',
the adversary lures the agent to a designated
target state by integrating a
generative model and a planning algorithm.
The generative model is used for predicting the future states of the agent, whereas the planning algorithm generates the actions for luring it. The attacks are successfully tested against the agents trained by the state-of-the-art deep reinforcement learning algorithms \cite{Voloymer_Human-level}, \cite{Volodymyr_Async}. Details on this work and example videos of the adversarial attacks can be found on the following URL: \url{http:
//yclin.me/adversarial_attack_RL/}.

In another work, Huang et al.~\cite{Huang_2017a} demonstrated that FGSM~\cite{Goodfellow_2015} can also be  used to significantly degrade performance
of trained policies in the context of deep reinforcement learning. Their  threat model considers adversaries that are capable of introducing
minor perturbations to the raw input of the policy.
The conducted experiments demonstrate that it is fairly easy to confuse neural network policies with adversarial examples, even in black-box scenarios.
Videos and further details on this work are available on \url{http://rll.berkeley.edu/adversarial/}.

\subsubsection{Attacks on Semantic Segmentation and Object Detection}
Semantic image segmentation and object detection are among the mainstream problems in Computer Vision. Inspired by Moosavi-Dezfooli~\cite{Uni}, Metzen et al.~\cite{Metzen} showed the existence of image-agnostic quasi-imperceptible  perturbations that can fool a deep neural network into significantly corrupting the predicted segmentation of the images. Moreover, they also showed that it is possible to compute noise vectors  that can remove a specific class from the segmented classes while keeping most of the image segmentation unchanged (e.g.~removing pedestrians from road scenes). Although it is argued that the ``space of the adversarial perturbations for the semantic image segmentation is presumably smaller than image classification'', the perturbations have been shown to generalize well for unseen validation images with high probability. 
Arnab et al.~\cite{Arnab_2017} also evaluated FGSM~\cite{Goodfellow_2015} based adversarial attacks for semantic segmentation and noted that many observations about these attacks for classification do not directly transfer to segmentation task.

\begin{figure}[t] %  figure placement: here, top, bottom, or page
   \centering
   \includegraphics[width=2.8in]{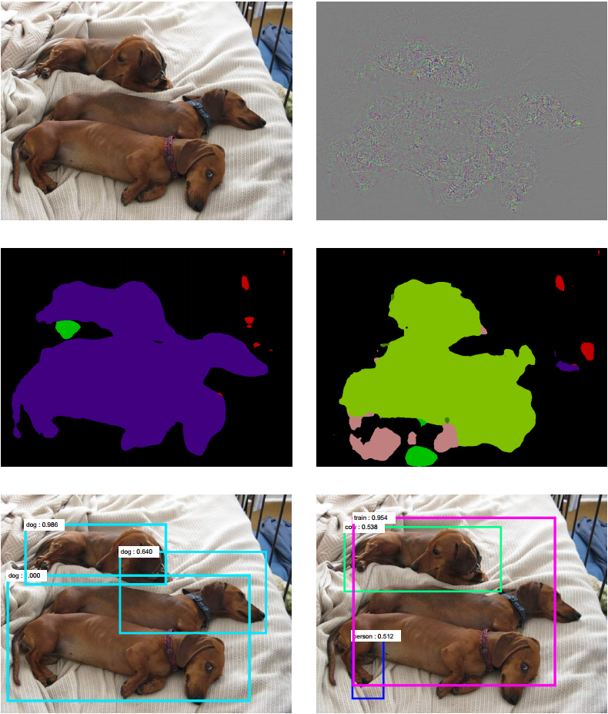} 
   \caption{Adversarial example for semantic segmentation
and object detection~\cite{Xie_2017}. FCN~\cite{FCN} and Faster-RCNN~\cite{FasterRCNN} are used for segmentation
and  detection, respectively.
Left column (top-down): Clean image, normal
segmentation (purple region is predicted as dog)
and detection results. Right column (top-down): Perturbation 10x, fooled segmentation (light green region is predicted as 
train and the pink region as person) and detection results.}  
   \label{fig:sement}
\end{figure}

Xie et al.~\cite{Xie_2017} computed adversarial examples for semantic segmentation and object detection under the observation that these tasks can be formulated as  classifying multiple targets in an image - the target is a pixel or a receptive field in segmentation, and object proposal in detection. Under this perspective, their approach, called `Dense Adversary Generation' optimizes a loss function over a set of pixels/proposals to generate adversarial examples. The generated examples are tested to fool a variety of deep learning based segmentation and detection approaches. Their experimental evaluation not only demonstrates successful fooling of the targeted networks but also shows that the generated perturbations generalize well across different network models. In Fig.~\ref{fig:sement}, we show a representative example of network fooling for segmentation and detection using the approach in \cite{Xie_2017}.

\begin{figure}[t] %  figure placement: here, top, bottom, or page
   \centering
   \includegraphics[width=2.8in]{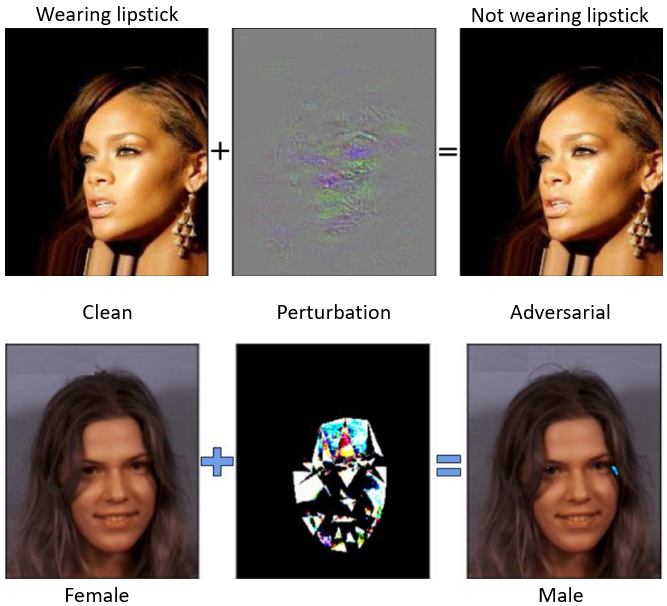} 
   \caption{Top-row: Example of changing a facial attribute `wearing lipstick' to `not wearing lipstick' by Fast Flipping Attribute method~\cite{Rozsa_Face}. Bottom row: Changing gender with perturbation generated by \cite{Face_Gender}.} 
   \label{fig:face}
\end{figure}

\section{Attacks in the real world}
\label{sec:Real}
\begin{figure*}[t] %  figure placement: here, top, bottom, or page
   \centering
   \includegraphics[width=5.3in]{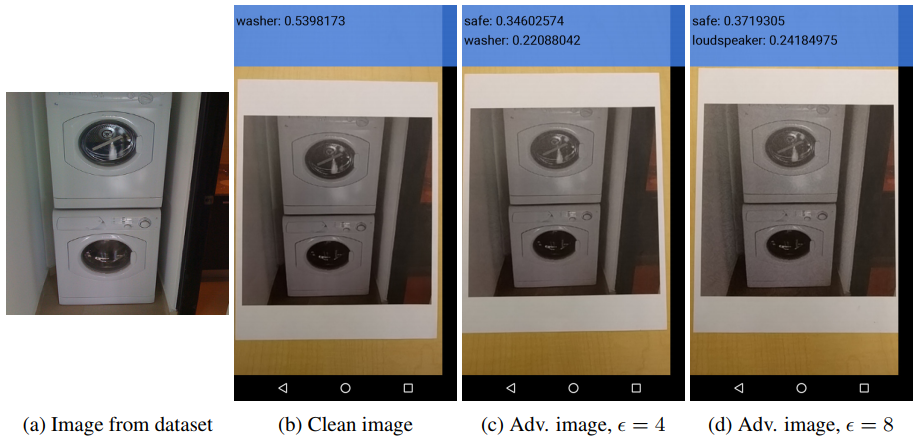} 
   \caption{Example of adversarial attack on mobile phone cameras: A clean image (a) was taken and used to generate different adversarial images. The images were printed and the TensorFlow
Camera Demo app~\cite{Mobileapp} was used to classify them. A clean image (b) is recognized correctly as a `washer' when
perceived through the camera, whereas adversarial images (c) and (d) are mis-classified. The images also show network confidence in the range [0,1] for each image. The value of $\epsilon$ is given for (\ref{eq:FGSM}). }  
   \label{fig:real}
\end{figure*}

\subsubsection{Attacks on Face Attributes}
Face attributes are among the emerging soft biometrics for modern security systems. Although face attribute recognition can also be categorized as a classification problem,  we separately review some interesting attacks in this direction because face recognition itself is treated as a mainstream problem in Computer Vision. 

Rozsa et al.~\cite{Rozsa_Face}, \cite{Rozsa_Are} explored the stability of multiple deep learning approaches using the CelebA benchmark~\cite{CelebA} by generating adversarial examples to alter the results of facial attribute recognition, see top-row in Fig.~\ref{fig:face}. By attacking the deep network classifiers with their so-called `Fast Flipping Attribute' technique, they found that robustness of deep neural networks against the adversarial attacks varies highly between facial attributes. It is claimed that adversarial attacks are very effective in changing the label of a target attribute to a correlated attribute. 
Mirjalili and  Ross~\cite{Face_Gender} proposed a
technique that modifies a face image such that its gender
(for a gender classifier) is modified, whereas its
biometric utility for a face matching system remains intact, see bottom-row in Fig.~\ref{fig:face}.
Similarly, Shen et al.~\cite{Shen_2017attractive} proposed two different techniques to generate adversarial examples for faces that can have high `attractiveness scores' but low `subjective scores' for the face attractiveness evaluation using deep neural network. We refer to \cite{FaceCrime} for further attacks related to the task of face recognition.

The literature reviewed in Section~\ref{sec:Attacks} assumes settings where adversaries directly feed deep neural networks with  perturbed images. Moreover, the effectiveness of attacks are also evaluated using standard image databases. Whereas those settings have proven sufficient to convince many researchers that adversarial attacks are a real concern for deep learning in practice, we also come across instances in the literature (e.g.~\cite{Graese_2016}, \cite{Lu_2017}) where this concern is down-played and adversarial examples are implicated to be `only a matter of curiosity' with little practical concerns. 
Therefore, this Section is specifically  dedicated to the literature that deals with the adversarial attacks in practical real-world conditions to help settle the debate.

\subsection{Cell-phone camera attack}
Kurakin et al.~\cite{Kurakin_2016a} first demonstrated that threats of adversarial attacks also exist in the physical world. To illustrate this, they printed adversarial images and took  snapshots from a cell-phone camera. These images were fed to TensorFlow Camera Demo app~\cite{Mobileapp} that uses Google's Inception model~\cite{Inceptionv3} for object classification. It was shown  that a large fraction of images were misclassified even when perceived through the camera. In Fig.~\ref{fig:real}, an example is shown from the original paper. A video is also provided on the following URL \url{https://youtu.be/zQ_uMenoBCk} that shows the threat of adversarial attacks with further images.
This work studies FGSM~\cite{Goodfellow_2015}, BIM and ILCM~\cite{Kurakin_2016a} methods for attacks in the  physical world.

\begin{figure*}[t] %  figure placement: here, top, bottom, or page
   \centering
   \includegraphics[width=5.7in]{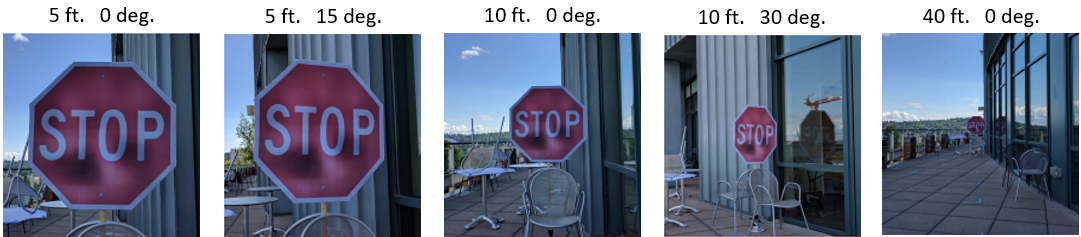} 
   \caption{Example of road sign attack~\cite{Evtimov_2017}: The success rate of fooling LISA-CNN~\cite{Evtimov_2017} classifier on all the shown images is $100\%$. The distance and angle to the camera are also shown. The classifier is trained using LISA dataset for road signs~\cite{LISA}.}  
   \label{fig:stop}
\end{figure*}

\subsection{Road sign attack}
Etimov et al.~\cite{Evtimov_2017} built on the attacks proposed in \cite{Carlini_2016} and \cite{Liu_2017b} to design robust perturbations for the physical world. They demonstrated the possibility of attacks that are robust to physical conditions, such as variation in view angles, distance and resolution.
The proposed algorithm, termed RP$_2$ for Robust Physical Perturbations, was used to generate adversarial examples for road sign recognition systems that achieved high fooling ratios in practical drive-by settings.
Two attack classes were introduced in this work for the physical road signs, (a) poster-printing: where the attacker prints a perturbed road sign poster and places it over the real sign (see Fig.~\ref{fig:stop}), (b) sticker perturbation: where the printing is done on a paper and the paper is stuck over the real sign. For (b) two types of perturbations were studied, (b1) subtle perturbations: that occupied the entire sign and (b2) camouflage perturbations: that took the form of graffiti sticker  on the sign. As such, all these perturbations require access to a color printer and no other special hardware.
Successful generation of perturbations for both (a) and (b) such that the perturbations remained robust to natural variations in the physical world demonstrate the threat of adversarial example in the real world.
We refer to the following URL for further details and videos related to this work: \url{https://iotsecurity.eecs.umich.edu/#roadsigns}.

It should be noted that Lu et al.~\cite{Lu_2017} had previously claimed that adversarial examples are not a concern for object detection in Autonomous Vehicles because of the changing physical conditions in a moving car. However, the attacking methods they employed \cite{Szegedy_2014}, \cite{Goodfellow_2015}, \cite{Kurakin_2016a} were somewhat primitive. The findings of Etimov et al.~\cite{Evtimov_2017} are orthogonal to the results in~\cite{Lu_2017b}.
However, in a follow-up work Lu et al.~\cite{Lu_2017a} showed that the detectors like YOLO 9000~\cite{YOLO9000} and FasterRCNN~\cite{FasterRCNN} are `currently' not fooled by the attacks introduced by Etimov et al.~\cite{Evtimov_2017}. 
Zeng et al.~\cite{Zeng_2017} also argue that adversarial perturbations in the image space do not generalize well in the physical space of the real-world. However, Athalye et al.~\cite{Athalye_2017} showed that we can actually print 3D physical objects for successful adversarial attacks in the physical world. We discuss~\cite{Athalye_2017} in Section~\ref{sec:3d}.   

Gu et al.~\cite{Gu_2017} also explored an interesting notion of threats to outsourced training of the neural networks in the context of fooling neural networks on street signs. They showed that it is possible to train a network (a \emph{BadNet}) that shows state-of-the-art performance on the user's training and validation samples, but behaves badly on attacker-chosen inputs. They demonstrated this attack in a realistic scenario by creating a street sign classifier that identifies stop signs as speed limits when a special sticker is added to the stop sign. Moreover, it was found that the fooling of the network persisted to a reasonable extent even when the network was later fine-tuned with additional training data.

\subsection{Generic adversarial 3D objects}
\label{sec:3d}
\begin{figure*}[t] %  figure placement: here, top, bottom, or page
   \centering
   \includegraphics[width=5.1in]{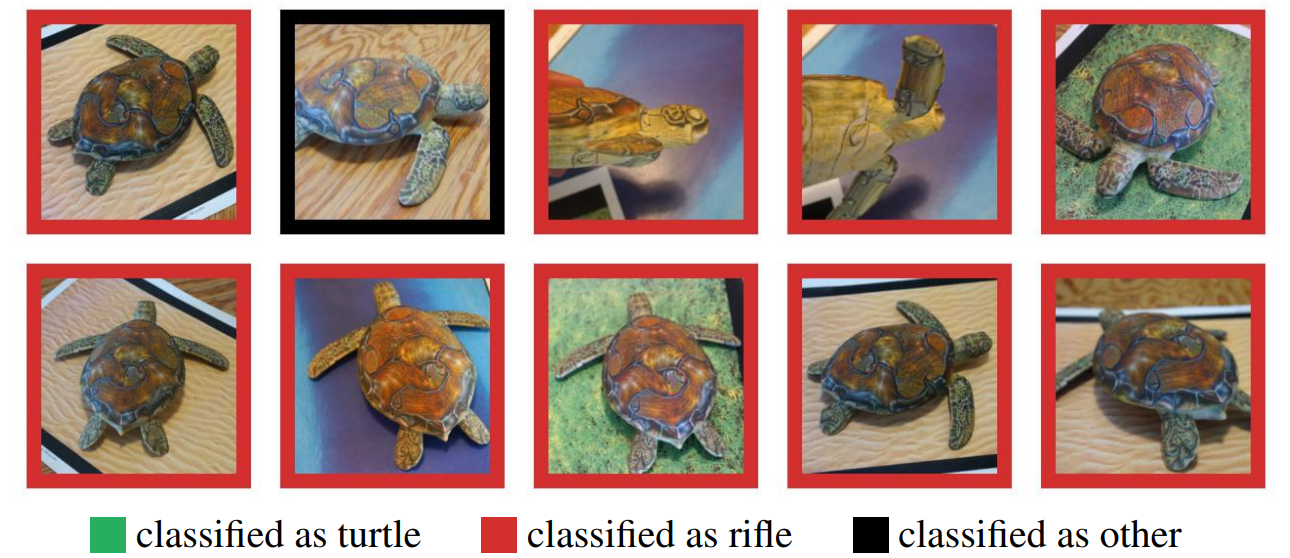} 
   \caption{Different random poses of a 3D-printed turtle perturbed by EOT \cite{Athalye_2017} to be classified as a rifle by an ImageNet classifier. The unperturbed version (not shown) is classified correctly with $100\%$ probability.}  
   \label{fig:3D}
\end{figure*}

Athalye et al.~\cite{Athalye_2017} introduced a method for constructing 3D objects that can fool neural networks across a wide variety of angles and viewpoints. Their `Expectation Over Transformation' (EOT) framework is able to construct examples that are adversarial over an entire distribution of image/object transformations. Their end-to-end approach is able to print arbitrary adversarial 3D objects. In our opinion, results of this work ascertain that adversarial attacks are a real concern for deep learning in the physical world. In Fig.~\ref{fig:3D} we show an example of 3D-printed turtle that is modified by EOT framework to be classified as rifle.
A video demonstrating the fooling by EOT in  the physical world is available at the following URL: \url{https://www.youtube.com/watch?v=YXy6oX1iNoA&feature=youtu.be}.

\subsection{Cyberspace attacks}
Papernot et al.~\cite{Papernot_2017b} launched one of the first attacks against the deep neural network classifiers in  cyberspace in the real-world settings. They trained a substitute network for the targeted black-box classifier on synthetic data, and instantiated the attack against remotely hosted neural networks by MetaMind, Amazon and Google.
They were able to show that the respective targeted networks misclassified $84.24 \%$, $96.19\%$ and $88.94\%$ of the adversarial examples generated by their method.
Indeed, the only information available to the attacker in their threat model was the output label of the targeted network for the input image fed by the attacker.
In a related work, Liu et al.~\cite{Liu_2017b} developed an ensemble based attack and showed its success against Clarifai.com - a commercial company providing state-of-the-art image classification services. The authors claim that their attacks for both targeted and non-targeted fooling are able to achieve high success rates.

Grosse et al.~\cite{Grosse_2016b} showed construction of effective
adversarial attacks for neural networks used
as malware classifiers. As compared to image recognition, the domain of malware classification introduces additional constraints in the adversarial
settings, e.g. continuous input domains
are replaced by discrete inputs,  the  condition of visual similarity 
is replaced by requiring equivalent functional behavior. However,  Grosse et al.~\cite{Grosse_2016b} showed that creating effective adversarial examples is still possible for maleware classification. Further examples of successful adversarial attacks against deep lrearning based malware classification can also be found in \cite{Rosenberg_2017}, \cite{Grosse_malware},   \cite{Hu_2017a}. 

\subsection{Robotic Vision \& Visual QA Attacks}
Melis et al.~\cite{Melis_2017} demonstrated the vulnerability
of robots to the adversarial manipulations of the
input images using the techniques in \cite{Szegedy_2014}. The authors argue that strategies to enforce deep neural networks to learn more stable representations are necessary for secure robotics.
Xu et al.~\cite{Xu_2017} generated adversarial attacks for the Visual Turing Test, also known as `Visual Question Answer' (VQA). The authors show that the commonly used compositional and non-compositional VQA architectures that employ deep neural networks are vulnerable to adversarial attacks. Moreover, the adversarial examples are transferable between the models. They conclude that the ``adversarial examples pose real threats to not
only image classification models, but also more complicated
VQA models''~\cite{Melis_2017}.

\section{Existence of adversarial examples}
\label{sec:exist}
 In the literature related to adversarial attacks on deep learning in Computer Vision, there are varied views on the existence of adversarial examples. These views  generally align well with the local empirical observations made by the researchers while attacking or defending the deep neural networks.
However, they often fall short in terms of generalization. For instance, the popular linearity hypothesis of Goodfellow et al.~\cite{Goodfellow_2015} explains the FGSM and related attacks very well. However, Tanay and Griffin~\cite{Tanay_2016} demonstrated image classes that do not suffer from adversarial examples for linear classifier, which is not in-line with the linearity hypothesis. Not to mention, the linearity hypothesis itself deviates strongly from the previously prevailing opinion that the adversarial examples stem from highly non-linear decision boundaries induced by deep neural networks. There are also other examples in the literature where the linearity hypothesis is not directly supported~\cite{Luo_2015}.
 
Flatness of decision boundaries~\cite{Fawzi_2016}, large local curvature of the decision boundaries~\cite{Analysis} and low flexibility of the networks~\cite{OldAnalysis} are some more examples of the viewpoints on the existence of adversarial examples that do not  perfectly align with each other. Whereas it is apparent that adversarial examples can be formed by modifying as little as one pixel in an image, current literature seems to lack consensus on the reasons for the existence of the adversarial examples.
%, currently no single theory explains the existence of all kinds of adversarial examples in a unified manner. 
This fact also makes analysis of adversarial examples an active research direction that is expected to explore and explain the nature of the decision boundaries induced by deep neural networks, which are currently more commonly treated as black-box models. 
Below, we review the works that mainly focus on analyzing the existence of  adversarial perturbations for deep learning.
We note that, besides the literature reviewed below, works related to adversarial attacks (Section~\ref{sec:Attacks}) and defenses (Section~\ref{sec:Defense}) often   provide brief analysis of adversarial perturbations while  conjecturing about the phenomena resulting in the existence of the adversarial examples.

%Below, we restrict our review to only those works that are primarily concerned with the analysis of adversarial perturbations. \\

%\noindent{\bf Fundamental limits on adversarial robustness:} 
\subsection{Limits on adversarial robustness}
Fawzi et al.~\cite{Fawzi_2015c} introduced a framework for studying the instability of classifiers to adversarial perturbations. 
They established fundamental limits on the robustness of classifiers in terms of a `distinguishability measure' between the classes of the dataset, where distinguishability is defined as the distance between the means of two classes for linear classifiers and the  distance between the matrices of second order moments for the studied non-linear classifiers. This work shows that adversarial examples also exist for the classifiers beyond deep neural networks.
The presented analysis traces back the phenomenon of adversarial instability to the low flexibility of the classifiers, which is not completely orthogonal to the prevailing belief at that time that high-nonlinearity of the networks make them susceptible to adversarial examples.

\subsection{Space of adversarial examples} 
Tabacof and Eduardo~\cite{Tabacof_2015} generated adversarial examples for shallow and deep network classifiers on MNIST~\cite{LeCun1989} and ImageNet~\cite{ImageNet} datasets and probed the pixel space of adversarial examples by using noise of varying distribution and intensity. The authors empirically demonstrated that adversarial examples appear in large regions in the pixel space, which is in-line with the similar claim in \cite{Goodfellow_2015}. However, somewhat in contrast to the linearity hypothesis, they argue that a weak, shallow and more linear classifier is also as susceptible to adversarial examples as a strong deep classifier. 

Tramer et al.~\cite{Tramer_2017space} proposed a method to estimate the dimensionality of the space of the adversarial examples. It is claimed that the adversarial examples span a contiguous high dimension space (e.g.~with dimensionality $\approx 25$).
Due to high dimensionality, the subspaces of different classifiers can intersect, which gives rise to the transferability of the adversarial examples. Interestingly, their analysis suggests that it is possible to defend classifiers against transfer-based attacks even when they are vulnerable to direct attacks.

\subsection{ Boundary tilting perspective} 
Tanay and Griffin~\cite{Tanay_2016} provided a `boundary tilting' perspective on the existence of adversarial examples for deep neural networks.
They argued that generally a single class data that is sampled to learn and evaluate a classifier lives in a sub-manifold of the class, and adversarial examples for that class exist when the classification boundary lies close to this sub-manifold.
They formalized the notion of `adversarial strength' of a classifier and reduced it to the `deviation angle' between the boundaries of the considered classifier and the nearest centroid classifier.
It is then shown that adversarial strength of a classifier can be varied by decision `boundary tilting'. The authors also argued that adversarial stability of the classifier is associated with its  regularization.
In the opinion of Tanay and Griffin, the linearity hypothesis~\cite{Goodfellow_2015} about the existence of adversarial examples is ``unconvincing''.

\subsection{Prediction uncertainty and evolutionary stalling of training cause adversaries}
Cubuk et al.~\cite{Cubuk_2017a} argue that the ``origin of adversarial examples is primarily due to an inherent uncertainty that neural networks have about their predictions''.
They empirically compute a functional form of the uncertainty, which is shown to be independent of network architecture, training protocol and dataset. 
It is argued that this form only depends on the statistics of the network logit differences.
This eventually results in fooling ratios caused by adversarial attacks to exhibit a universal scaling with respect to the size of perturbation.
They studied FGSM~\cite{Goodfellow_2015}, ILCM and BIM~\cite{Kurakin_2016a}  based attacks to corroborate their claims. It is also claimed that accuracy of a network on clean images correlates with its adversarial robustness (see Section~\ref{sec:ARC} for more arguments in this direction).

Rozsa et al.~\cite{Rozsa_2017c} hypothesized that the existence of adversarial perturbations is a result of evolutionary stalling of decision boundaries on training images. In their opinion, individual training samples stop contributing to the training loss of the model (i.e. neural network) once they are classified correctly, which can eventually leave them close to the decision boundary. Hence, it becomes possible to throw those (and  similar) samples away to a wrong class region by adding minor perturbations.
They proposed a Batch Adjusted Network Gradients (BANG) algorithm to train a network to mitigate the evolutionary stalling during training. 

\subsection{Accuracy-adversarial robustness correlation}
\label{sec:ARC}
In the quest of explaining the existence of adversarial perturbations, Rozsa et al.~\cite{Rozsa_2016b} empirically analyzed the correlation between the accuracy of eight deep network classifiers and their robustness to three adversarial attacks introduced in~\cite{Goodfellow_2015},\cite{Rozsa_2016}.
The studied classifiers include AlexNet~\cite{Alex_2012}, VGG-16 and VGG-19 networks~\cite{VGG}, Berkeley-trained  version of GoogLeNet and Princeton-GoogLeNet~\cite{GoogleNet}, ResNet-52; ResNet-101; and ResNet-152~\cite{ResNet}. The adversarial examples are generated with the help of large-scale ImageNet dataset~\cite{ImageNet} using the techniques proposed in \cite{Goodfellow_2015} and \cite{Rozsa_2016}. Their experiments lead to the observation that the networks with higher classification accuracy generally also exhibit more robustness against the adversarial examples.
They also concluded that adversarial examples transfer better between similar network topologies.

\subsection{ More on linearity as the source}
Kortov and Hopfiled~\cite{Krotov_2017} examined the existence of adversarial perturbations in the context of Dense Associative Memory (DAM) models~\cite{DAM}. 
As compared to the typical modern deep neural networks, DAM models employ higher order (more than quadratic) interactions between the neurons. The authors have demonstrated that adversarial examples generated using DAM models with smaller interaction power, which is similar to using a deep neural network with ReLU activation \cite{ReLU} for inducing linearity, are unable to fool models having higher order interactions.
The authors provided empirical evidence on the existence of adversarial examples that is independent of the FGSM~\cite{Goodfellow_2015} attack, yet supports the linearity hypothesis of Goodfellow et al.~\cite{Goodfellow_2015}.

\subsection{Existence of universal perturbations}
Moosavi-Dezfooli et al.~\cite{Uni} initially argued that universal adversarial perturbations exploit geometric correlations between the decision boundaries induced by the classifiers. Their existence partly owes to a subspace containing normals to the decision boundaries, such that the normals also surround the natural images. In~\cite{Analysis}, they built further on their theory and showed the existence of common directions (shared across datapoints) along which the decision boundary of a classifier can be highly positively curved. They argue that such directions play a key role in the existence of universal perturbations.
Based on their findings, the authors also propose a new geometric method to efficiently compute universal adversarial perturbations. 

It is worth noting that previously Fawzi et al.~\cite{Fawzi_2016} also associated the theoretical bounds on the robustness of classifiers to the curvature of decision boundaries. Similarly, Tramer et al.~\cite{Tramer_2017} also held  the curvature of decision boundaries in the vicinity of data points responsible for the vulnerability of neural networks to black-box attacks. In another recent work, Mopuri et al.~\cite{NAG} present a GAN-like model to learn the distribution of the universal adversarial perturbations for a given target model. The learned distributions are also observed to show good transferability across models. \\

% \begin{enumerate}
% \item Fundamental limits on adversarial robustness \cite{Fawzi_2015c}
% \item Exploring the Space of Adversarial Images \cite{Tabacof_2015}
% \item The Space of Transferable Adversarial Examples \cite{Tramer_2017space}
% \item A Boundary Tilting Persepective on the Phenomenon of Adversarial Examples \cite{Tanay_2016}.
% \item Are Accuracy and Robustness Correlated? \cite{Rozsa_2016b}
% \item Analysis of universal adversarial perturbations \cite{Analysis}.
% \item Intriguing Properties of Adversarial Examples \cite{Cubuk_2017a}
% \item Dense Associative Memory is Robust to Adversarial Inputs \cite{Krotov_2017}
% \item Towards Robust Deep Neural Networks with BANG\cite{Rozsa_2017c}
% \end{enumerate}

\section{Defenses against adversarial attacks}
\label{sec:Defense}
%In this section, we review the approaches aimed at defending deep neural networks against the adversarial attacks. 
Currently, the defenses against the adversarial attacks are being developed along three main directions:

\begin{enumerate}
\item Using {\it modified training} during learning or {\it modified input} during testing.
\item {\it Modifying networks}, e.g.~by adding more layers/sub-networks, changing loss/activation functions etc.
\item Using external models as {\it network add-on} when classifying unseen examples. 
\end{enumerate}

\begin{figure}[t] %  figure placement: here, top, bottom, or page
   \centering
   \includegraphics[width=2.9in]{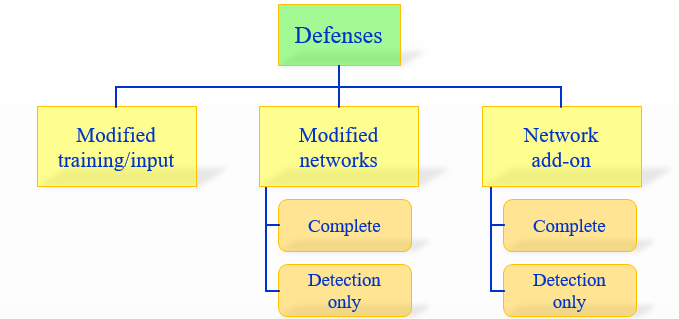} 
   \caption{Broad categorization of approaches aimed at defending deep neural networks against adversarial attacks.}  
   \label{fig:tax}
\end{figure}

The approaches along the first direction do not directly deal with the learning  models. On the other hand, the other two categories are more concerned with the  neural networks themselves. The techniques under these categories can be further divided into two types; namely (a) complete defense and (b) detection only.
The `complete defense' approaches aim at enabling the targeted network to achieve its original goal on the adversarial examples, e.g. a classifier predicting labels of adversarial examples with acceptable accuracy.
On the other hand, `detection only' approaches are meant to raise the red flag on potentially adversarial examples to reject them in any further processing. 
The taxonomy of the described categories is also shown in Fig.~\ref{fig:tax}.
The remaining section is organized according to this taxonomy. 
In the used taxonomy, the difference between `modifying' a network and employing an `add-on' is that the former makes changes to the original deep neural network architecture/parameters during training. On the other hand, the latter keeps the original model intact and appends external model(s) to it during testing.

\subsection{Modified training/input}
\subsubsection{ Brute-force adversarial training}
Since the discovery of adversarial examples for the deep neural networks~\cite{Szegedy_2014}, there has been a general consensus in the related literature that robustness of neural networks against these examples improves with adversarial training. Therefore, most of the contributions introducing new adversarial attacks, e.g.~\cite{Szegedy_2014},~\cite{Goodfellow_2015}, \cite{DeepFool} (see Section~\ref{sec:Attacks}) simultaneously propose adversarial training as the first line of defense against those attacks. 
Although adversarial training improves robustness of a network, to be really effective, it requires that training is performed using strong attacks and the architecture of the network is sufficiently expressive.
Since adversarial training necessitates increased training/data size, we refer to it as a `brute-force' strategy.

It is also commonly observed in the literature that brute-force adversarial training results in regularizing the network (e.g. see~\cite{Goodfellow_2015}, \cite{Sankaranarayanan_2017}) to reduce over-fitting, which in turn improves robustness of the networks against the adversarial attacks. Inspired by this observation, Miyato et al.~\cite{Miyato_2016} proposed a `Virtual Adversarial Training' approach to smooth the output distributions of the neural networks. A related `stability training' method is also proposed by Zheng et al.~\cite{Zheng_2016c} to improve the robustness of neural networks against small distortions to input images. It is noteworthy that whereas adversarial training is known to improve robustness of neural networks, Moosavi-Dezfooli~\cite{Uni} showed that effective adversarial examples can again be computed for already adversarially trained networks.

\subsubsection{Data compression as defense}
Dziugaite et al.~\cite{Dziugaite_2016} noted that most of the popular image classification datasets comprise JPG images. Motivated by this observation, they studied the effects of JPG compression on the perturbations computed by FGSM~\cite{Goodfellow_2015}. It is reported that JPG compression can actually reverse the drop in classification accuracy to a large extent for the FGSM perturbations. Nevertheless, it is concluded that compression alone is far from an effective defense. JPEG compression was also studied by Guo et al.~\cite{Guo_2017} for mitigating the effectiveness of adversarial images.  Das et al.~\cite{Das_2017} also took a similar approach and used JPEG compression to remove the high frequency components from images and proposed an ensemble-based technique to counter the adversarial attacks generated by FGSM~\cite{Goodfellow_2015} and DeepFool method~\cite{DeepFool}. Whereas encouraging results are reported in~\cite{Das_2017}, there is no analysis provided for the stronger attacks, e.g.~C\&W attacks~\cite{Carlini_2016}. Moreover, Shin and Song~\cite{JPEGdist} have demonstrated the existence of adversarial examples that can survive JPEG compression.
Compression under Discrete Cosine Transform (DCT) was also found  inadequate as a defense against the universal perturbations~\cite{Uni} in our previous work~\cite{Akhtar_2017}. One major limitation of compression based defense is that larger compressions also result in loss of classification  accuracy on clean images, whereas smaller compressions often do not adequately remove the adversarial perturbations.  

In another related approach, Bhagoji et al.~\cite{Bhagoji_2017} proposed to compress input data using Principal Component Analysis for adversarial robustness. However, Xu et al.~\cite{Xu_2017Squueze} noted that this compression also results in  corrupting the spatial structure of the image, hence often adversely affecting the classification performance.

\subsubsection{Foveation based defense}
Luo et al.~\cite{Luo_2015} demonstrated that significant robustness against the adversarial attacks using L-BFGS~\cite{Szegedy_2014} and FGSM~\cite{Goodfellow_2015} is possible with `foveation' mechanism - applying neural network in different regions of images. It is hypothesized that CNN-based classifiers trained on large datasets, such as ImageNet~\cite{ImageNet} are generally robust to scale and translation variations of objects in the images. However, this invariance does not extend to  adversarial patterns in the images. This makes foveation as a viable option for reducing the effectiveness of adversarial attacks proposed in \cite{Szegedy_2014}, \cite{Goodfellow_2015}. However, foveation is yet to demonstrate its effectiveness against  more powerful attacks.

\subsubsection{Data randomization and other methods}
Xie et al.~\cite{Xie_2017} showed that random resizing of the adversarial examples reduces their effectiveness. Moreover, adding random padding to such examples also results in reducing the fooling rates of the networks.  Wang et al.~\cite{Wang_2016resistant} transformed the input data with a separate data-transformation module to remove possible adversarial perturbations in images.  In the literature, we also find evidence that data augmentation during training (e.g.~Gaussian data augmentation~\cite{Zantedeschi_2017}) also helps in improving robustness of neural networks to adversarial attacks, albeit only slightly.

\subsection{Modifying the network}
\label{sec:ModNet}
For the approaches that modify the neural networks for defense against the adversarial attacks, we first discuss the `complete defense' approaches.  
The `detection only' approaches are separately reviewed in Section~\ref{sec:detect_network}.

\subsubsection{Deep Contractive Networks}
In the early attempts of making deep learning robust to adversarial attacks, Gu and Rigazio~\cite{Gu_2015}  introduced Deep Contractive Networks (DCN). It was shown that Denoising Auto Encoders~\cite{AE} can reduce adversarial noise, however stacking them with the original networks can make the resulting network even more vulnerable to perturbations. Based on this observation, the training procedure of DCNs used a smoothness penalty similar to Contractive Auto Encoders~\cite{CAE}. Whereas reasonable robustness of DCNs was demonstrated against the L-BGFS~\cite{Szegedy_2014} based attacks, many stronger attacks have been introduced since DCNs were initially proposed. A related concept of using auto encoders for adversarial robustness of the neural networks can be also found in \cite{Bai_2017auto}. 

\subsubsection{Gradient regularization/masking}
Ross and Doshi-Velez~\cite{Ross_2017} studied  input gradient regularization~\cite{Druker_92} as a method for adversarial robustness. Their method trains differentiable models (e.g.~deep neural networks) while penalizing the degree of variation resulting in the output with respect to change in the input. Implying, a small adversarial  perturbation becomes unlikely to change the output of the trained model drastically. It is shown that this method, when combined with brute-force adversarial training, can result in very good robustness against attacks like FGSM~\cite{Goodfellow_2015} and JSMA~\cite{Papernot_2016c}. However, each of these methods almost double the training complexity of a network, which is already prohibitive in many cases. 
%We note that, this is also the case for the defensive distillation method~\cite{Papernot_2016} discussed above.

Previously, Lyu et al.~\cite{Lyu_2015} also used the notion of penalizing the gradient of loss function of network models with respect to the inputs to incorporate robustness in the networks against L-BFGS~\cite{Szegedy_2014} and FGSM~\cite{Goodfellow_2015} based attacks. Similarly, Shaham et al.~\cite{Shaham_2016} attempted to improve the local stability of neural networks by minimizing the loss of a model over adversarial examples at each parameter update.
They minimized the loss of their model over worst-case adversarial examples instead of the original data. In a related work, Nguyen and Sinha~\cite{Nguyen_2017}  introduced a masking based defense against C\&W attack~\cite{Carlini_2016} by adding noise to the logit outputs of  networks. 

\subsubsection{Defensive distillation} Papernot et al.~\cite{Papernot_2016} exploited the notion of `distillation'~\cite{Hinton_2014} to make deep neural networks robust against adversarial attacks. Distillation was introduced by Hinton et al.~\cite{Hinton_2014} as a training procedure to transfer knowledge of a more complex network to a smaller network. The variant of the procedure introduced by Papernot et al.~\cite{Papernot_2016} essentially uses the knowledge of the network to improve its own robustness. The knowledge is extracted in the form of class probability vectors of the training data and it is fed back to train the original model. It is shown that doing so improves resilience of a network to small perturbation in the images. Further empirical evidence in this regard is also provided in~\cite{Papernot_2016On}. Moreover, in a follow-up work,  Papernot et al.~\cite{Papernot_extending} also extended the defensive distillation method by addressing the numerical instabilities that were encountered in \cite{Papernot_2016}. It is worth noting that the `Carlini and Wagner' (C\&W) attacks~\cite{Carlini_2016} introduced in Section~\ref{sec:Popular} are claimed to be successful against the defensive distillation technique. We also note that defensive distillation can also be seen as an example of `gradient masking' technique. However, we describe it separately keeping in view its popularity in the literature. 

\subsubsection{Biologically inspired protection}
Nayebi and Ganguli~\cite{Nayebi_2017} demonstrated natural robustness of neural networks against adversarial attacks with highly non-linear activations (similar to nonlinear dendritic computations). It is noteworthy that the Dense Associative Memory models of Krotov and Hopfield~\cite{Krotov_2017} also work on a similar principle for robustness against the adversarial examples. Considering the linearity hypothesis of Goodfellow et al.~\cite{Goodfellow_2015}, \cite{Nayebi_2017} and \cite{Krotov_2017} seem to further the notion of susceptibility of  modern neural networks to adversarial examples being the effect of linearity of activations. 
We note that Brendel and Bethge~\cite{BrokenBio} claim that the attacks fail on the biologically inspired protection~\cite{Nayebi_2017} due to numerical limitations of  computations. Stabilizing the computations again allow successful attacks on the  protected networks.

\subsubsection{Parseval Networks}  Cisse et al.~\cite{Cisse_2017} proposed `Parseval' networks as a defense against the adversarial attacks. These networks employ a layer-wise regularization by controlling the global Lipschitz constant of the network. Considering that a network can be seen as a composition of functions (at each layer), robustification against small input perturbations is possible by maintaining a small Lipschitz constant for these functions. Cisse et al. proposed to do so by controlling  the spectral norm of the weight matrices of the networks by parameterizing them with `parseval tight frames' \cite{Parse}, hence the name `Parseval' networks.

\subsubsection{DeepCloak}
Gao et al.~\cite{Gao_2017DeepCloak} proposed to insert a masking layer immediately before the layer handling the classification. The added layer is explicitly trained by forward-passing clean and adversarial pair of images, and it encodes the differences between the output features of the previous layers for those image pairs. It is argued that the most dominant weights in the added layer correspond to the most sensitive features of the network (in terms of adversarial manipulation).  Therefore, while classifying, those features are masked by forcing the dominant weights of the added layer to zero.

\subsubsection{Miscellaneous approaches}
Among other notable efforts in making neural networks robust to adversarial attacks, Zantedeschi et al.~\cite{Zantedeschi_2017} proposed to use bounded ReLU~\cite{BRelu} to reduce the effectiveness of adversarial patterns in the images. Jin et al.~\cite{Jin_2015} introduced a feedforward CNN that used additive noise to mitigate the effects of adversarial examples. 
Sun et al.~\cite{Sun_2017a} prposed `HyperNetworks' that use statistical filtering as a method to make the network robust. 
Madry et al.~\cite{Madry_2017} studied adversarial defense from the perspective of robust optimization.
They showed that adversarial training with a PGD adversary can successfully defend against a range of other adversaries.
Later,  Carlini et al.~\cite{Carlini_2017b} also verified this observation.
Na et al.~\cite{Na_2017} employed a network that is regularized with a unified embedding for classification and low-level similarity learning.
The network is penalized using the distance between  clean and the  corresponding adversarial embeddings.
Strauss et al.~\cite{Strauss_2017} studied ensemble of methods to defend a network against the perturbations. 
Kadran et al.~\cite{Kardan_2017} modified the output layer of a neural network to induce robustness against the adversarial attacks. Wang et al.~\cite{Wang_2016a}, \cite{Wang_2016} developed adversary resistant neural networks by leveraging non-invertible data transformation in the network.  
Lee et al.~\cite{Lee_2016c} developed manifold regularized networks that use a training objective to minimizes the difference between multi-layer embedding results of clean and adversarial images.
Kotler and Wong~\cite{Kolter_2017} proposed to learn ReLU-based classifier that show robustness against small adversarial perturbations.
They train a neural network that provably achieves high accuracy (>90\%) against any adversary in a canonical setting ($\epsilon=0.1$ for $\ell_{\infty}$-norm perturbation on MNIST).
Raghunathan et al.~\cite{Certified} studied the problem of defense for neural networks with one hidden layer. Their approach produces a network and a certificate on MNIST dataset such that no attack perturbing image pixels by at most $\epsilon = 0.1$ could results in more than 35\% test error.
%However, given that their approach is computationally infeasible to apply on larger networks, the only defenses that have been extensively evaluated are those of Madry et al.~\cite{Madry_2017}, giving 89\% accuracy against large $\epsilon$ (0.3/1) on MNIST and 45\% for moderate $\epsilon$ (8/255) on CIFAR for the $\ell_{\infty}$-norm perturbations.
Kolter and Wong~\cite{Kolter_2017} and Raghunathan et al.~\cite{Certified}  are among very few provable methods in defense against the adversarial attacks. Given that these methods are computationally infeasible to apply on larger networks, the only defenses that have been extensively evaluated are those of Madry et al.~\cite{Madry_2017} giving ~89\% accuracy against large epsilon (0.3/1) on MNIST and 45\% for moderate epsilon (8/255) on CIFAR. Another thread of works that can be seen as adversarial attacks/defenses with guarantees is related to verification of deep neural networks, e.g. \cite{Huang1}, \cite{Huang2}. 
In their approach OrOrbia et al.~\cite{DataGard} show that many different  proposals of adversarial training are instances of more general regularized objective, they termed DataGrad. The proposed DataGrad framework can be seen as an extension of layerwise contractive autoencoder penalty.

\subsubsection{Detection Only approaches}
\label{sec:detect_network}
\noindent{\bf SafetyNet}: Lu et al.~\cite{Lu_2017b} hypothesized that adversarial examples produce different patterns of ReLU activations in (the late stages of) networks than what is produced by  clean images.
Based on this hypothesis, they proposed to append a Radial Basis Function SVM classifier to the targeted models such that the SVM uses discrete codes computed by the late stage ReLUs of the network. 
To detect perturbation in a test image, its code is compared against those of training samples using the SVM.
Effective detection of adversarial examples generated by \cite{Goodfellow_2015},  \cite{Kurakin_2016a}, \cite{DeepFool} is demonstrated by their framework, named SafetyNet.\\

\noindent{\bf Detector subnetwork}:  Metzen et al.~\cite{Metzen_Ondetect} proposed to augment a targeted network with a subnetwork that is trained for a binary classification task of detecting adversarial perturbations in inputs.
It is shown that appending such a network to the internal layers of a model and using adversarial training can help in detecting perturbations generated using FGSM~\cite{Goodfellow_2015}, BIM~\cite{Kurakin_2016a} and DeepFool~\cite{DeepFool} methods. However, Lu et al.~\cite{Lu_2017b} later showed that this approach is again vulnerable to counter-counter measures.\\

\noindent{\bf Exploiting convolution filter statistics}: Li and Li~\cite{Li_ICCV17} used statistics of the convolution filters in CNN-based neural networks to classify  the input images as clean or adversarial. A cascaded classifier is designed that uses these statistics, and it is shown to detect more than $85\%$ adversarial images generated by the methods in \cite{Szegedy_2014}, \cite{Nguyen_2015}.
\\

\noindent{\bf Additional class augmentation}:
Grosse et al.~\cite{Grosse_2017} proposed to augment the potentially targeted neural network model with an additional class in which the model is trained to classify all the adversarial examples. Hosseini et al.~\cite{Hosseini_2017} also employed a similar strategy to detect black-box attacks.  

\subsection{Network add-ons}
\label{sec:Addons}
\subsubsection{Defense against universal perturbations}
\begin{figure}[t] %  figure placement: here, top, bottom, or page
   \centering
   \includegraphics[width=3.3in]{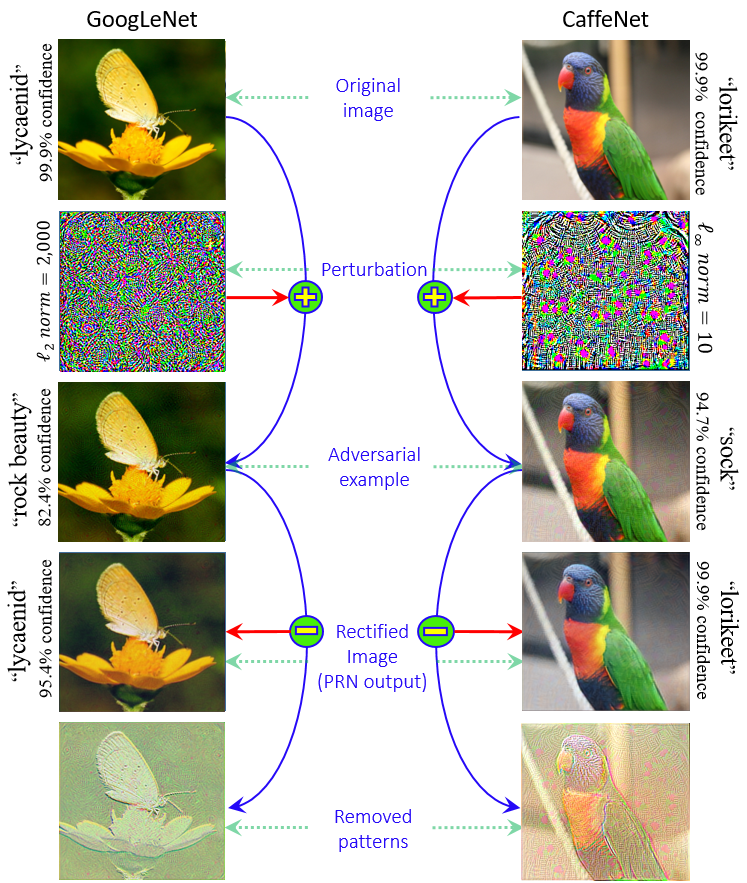} 
   \caption{Illustration of defense against universal perturbations~\cite{Akhtar_2017}: The approach rectifies an image to restore the network prediction. The pattern removed by rectification is separately analyzed to detect perturbation.}  
   \label{fig:UniDef}
\end{figure}

Akhtar et al.~\cite{Akhtar_2017} proposed a defense framework against the adversarial attacks generated using universal perturbations~\cite{Uni}. 
The framework appends extra `pre-input' layers to the targeted network and trains them to rectify a perturbed image so that the classifier's prediction becomes the same as its prediction on the clean version of the same image. The pre-input layers are termed Perturbation Rectifying Network (PRN), and they are trained without updating the parameters of the targeted network. A separate detector is trained by extracting features from the input-output differences of PRN for the training images. A test image is first passed through the PRN and then its features are used to detect perturbations. If adversarial perturbations are detected, the output of PRN is used to classify the test image. Fig.~\ref{fig:UniDef}, illustrates the rectification performed by PRN. The removed patterns are separately analyzed for detection.

\subsubsection{GAN-based defense}
Lee et al.~\cite{GANbased} used the popular framework of Generative Adversarial Networks~\cite{GANs} to train a network that is robust to FGSM~\cite{Goodfellow_2015} like attacks. The authors proposed to directly train the network along a generator network that attempts to generate perturbation for that network. During its training, the classifier keeps trying to correctly classify both the clean and perturbed images. We categorize this technique as an `add-on' approach because the authors propose to always train any network in this fashion. In another GAN-based defense, Shen et al.~\cite{withGAN} use the generator part of the network to rectify a perturbed image.

\subsubsection{Detection Only approaches}
\label{sec:detectors}
\noindent{\bf Feature squeezing}: Xu et al.~\cite{Squeez} proposed to  use feature squeezing to detect adversarial perturbation to an image.
They added two external models to the classifier network, such that these models reduce the color bit depth of each pixel in the image, and perform spatial smoothing over the image.
The predictions of the targeted network over the original image and the squeezed images are compared. If a large difference is found between the predictions, the image is considered to be an adversarial example. 
Whereas~\cite{Squeez} demonstrated the effectiveness of this approach against more classical attacks~\cite{Goodfellow_2015}, a follow-up work \cite{Xu_2017Squueze}   also claims that the method works reasonably well against the more powerful  C\&W attacks~\cite{Carlini_2016}. 
He et al.~\cite{He_2017} also combined feature squeezing with the ensemble method proposed in \cite{Abbasi_2017} to show that strength of defenses does  not always increase by combining them.
\\

\noindent{\bf MagNet}: Meng and Chen~\cite{Magnet} proposed a framework that uses one or more external detectors to classify an input image as adversarial or clean. During training, the framework aims at learning the manifold of clean images. In the testing phase, the images that are found far from the manifold are treated as adversarial and are rejected. The images that are close to the manifold (but not exactly on it) are always reformed to lie on the manifold and the classifier is fed with the reformed images.
The notion of attracting nearby images to the manifold of clean images and dropping the far-off images also inspires the name of the framework, i.e. MagNet. It is noteworthy that Carlini and Wagner~\cite{MagNetBreak} very recently demonstrated that this defense technique can also be defeated with slightly larger perturbations.\\

\noindent{\bf Miscellaneous methods}:
Liang et al.~\cite{Liang_2017} treated perturbations to images as noise and used scalar quantization and spatial smoothing filter to separately detect such perturbations. 
In a related approach, Feinman et al.~\cite{Feinman_2017} proposed to detect adversarial perturbations by harnessing uncertainty estimates (of dropout neural networks) and performing density estimation in the feature space of neural networks. Eventually, separate binary classifiers are trained as adversarial example detectors using the proposed features. 
Gebhart and Schrater~\cite{Gebhart_2017} viewed neural network computation
as information flow in graphs and proposed a method to detect adversarial perturbations by applying persistent homology to the induced graphs.

\section{Outlook of the research direction }
\label{sec:future}
In the previous sections, we presented a comprehensive review of the recent literature in adversarial attacks on deep learning. Whereas several interesting facts were reported in those sections along the technical details,  below we  make more general observations regarding this emerging research direction.
The discussion  presents a broader outlook to the readers without in-depth technical knowledge of this area. Our arguments are based on the literature reviewed above. \\ 
%The presented view-point  the  influential works in the  existing literature.\\

\noindent{\bf The threat is real:}  Whereas few works suggest that adversarial attacks on deep learning may not be a serious concern, a large body of the related literature indicates otherwise. The literature reviewed in Sections~\ref{sec:Attacks} and \ref{sec:Real} clearly demonstrate that  adversarial attacks can severely degrade the performance of deep learning techniques on multiple Computer Vision tasks, and beyond. In particular, the literature reviewed in Section~\ref{sec:Real} ascertains that deep learning is  vulnerable to  adversarial attacks in the real physical world. Therefore, we can conclusively argue that adversarial attacks pose a real threat to deep learning in practice. \\

\noindent{\bf Adversarial vulnerability is a general phenomenon:}  The reviewed literature shows successful fooling of different types of deep neural networks, e.g. MLPs, CNNs, RNNs on a variety of tasks in Computer Vision, e.g.~recognition, segmentation, detection. Although most of the existing works focus on fooling  deep learning on the task of classification/recognition, based on the surveyed literature we can easily observe that deep learning approaches are vulnerable to adversarial attacks in general. \\

\noindent{\bf Adversarial examples often generalize well:}  One of the most common properties of adversarial examples reported in the literature is that they transfer well between different neural networks. This is especially true for the networks that have relatively similar architecture.  The generalization of adversarial examples  is often exploited in black-box attacks.\\

\noindent{\bf Reasons of adversarial vulnerability need more investigation:} There are varied view-points in the literature on the reasons behind the vulnerability of deep neural networks to subtle adversarial perturbations. Often, these view-points are not well-aligned with each other. There is an obvious need for systematic investigation in this direction. \\

\noindent{\bf Linearity does promote vulnerability:} Goodfellow et al.~\cite{Goodfellow_2015} first suggested that the design of modern deep neural networks that forces them to behave linearly in high dimensional spaces also makes them vulnerable to adversarial attacks. Although popular, this notion  has also faced some opposition in the literature. Our  survey pointed out  multiple independent contributions that  hold linearity of the neural networks accountable for their vulnerability to adversarial attacks. Based on this fact, we can argue that linearity does promote vulnerability of deep neural networks to the adversarial attacks. However, it does not seem to be the only reason behind successful fooling of deep neural networks with cheap analytical perturbations. \\

\noindent{\bf Counter-counter measures are possible:}  Whereas multiple  defense techniques exist to counter  adversarial attacks, it is often shown in the literature that a defended model can again be successfully attacked by devising counter-counter measures, e.g. see \cite{Ensemble}.  This observation necessitates that new defenses also provide an estimate of their robustness against obvious counter-counter measures. \\

\noindent{\bf Highly active research direction:} The profound implications of vulnerability of deep neural networks to adversarial perturbations have made research in adversarial attacks and their defenses highly active in recent time.  The majority of the literature reviewed in this survey surfaced in the last two years, and there is currently a continuous influx of contributions in this direction. On one hand,   techniques are being proposed to defend neural networks  against the known attacks, on the other; more and mote powerful attacks are being devised. Recently, a Kaggle competition was also organized for the defenses against the adversarial attacks (\url{https://www.kaggle.com/c/nips-2017-defense-against-adversarial-attack/}). It can be hoped that this high research activity will eventually result in making deep learning approaches robust enough to be used in safety and security critical applications in the real world.

% \begin{enumerate}
% \item Ground-Truth Adversarial Examples \cite{Carlini_2017b}.
% \item Measuring Neural Net Robustness with Constraints \cite{Bastani_2017}
% \item On the Robustness of Convolutional Neural Networks to Internal Architecture and Weight Perturbations \cite{Cheney_2017}.
% \item A Closer Look at Memorization in Deep Networks \cite{Arpit_2017a}
% \item Towards proving the adversarial robustness of deep neural networks~\cite{Katz_2017}
% \item Towards Interpretable Deep Neural Networks by Leveraging Adversarial Examples \cite{Dong_2017Leverage}
%\end{enumerate}

\section{Conclusion}
\label{sec:conc}
This article presented the first comprehensive survey in the direction of adversarial attacks on deep learning in Computer Vision. Despite the high accuracies of deep neural networks on a wide variety of Computer Vision tasks, they have been found vulnerable to subtle input perturbations that lead them to completely change their outputs. With deep learning at the heart of the  current advances in machine learning and artificial intelligence, this finding has resulted in numerous recent contributions that devise adversarial attacks and their defenses for deep learning. 
This article reviews these contributions, mainly focusing on the most influential and interesting works in the literature. From the reviewed literature, it is apparent that adversarial attacks are a real threat to deep learning in practice, especially in safety and security critical applications.
The existing literature demonstrates that currently deep learning can not only be effectively attacked in cyberspace but also in the physical world.  However, owing to the very high activity in this research direction it can be hoped that deep learning will be able to show considerable robustness against the adversarial attacks in the future.

% \section*{Acknowledgement}
% The authors acknowledge Nicholas Carlini for useful suggestions for improvements.  

% or


% \appendices

% % you can choose not to have a title for an appendix
% % if you want by leaving the argument blank
% \section{}

% \section*{Acknowledgment}

% The authors would like to thank..

\ifCLASSOPTIONcaptionsoff
  \newpage
\fi



%\bibliographystyle{IEEEtran}

%----------------------------------------------------------------------------------------

\balance

%--------------------------------------------------------------------------------------


\begin{IEEEbiography}[{\includegraphics[width=1in,height=1.25in,clip,keepaspectratio]{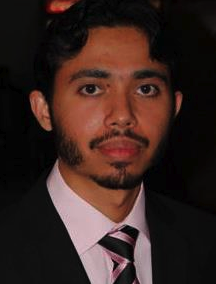}}]{Naveed Akhtar} %(M'76--SM'81--F'87) 
received his PhD in Computer Vision from The University of Western Australia (UWA) and Master degree in Computer Science from Hochschule Bonn-Rhein-Sieg, Germany (HBRS). His research in  Computer Vision and Pattern Recognition has been published in  prestigious venues of the field, including IEEE CVPR and IEEE TPAMI. He has also served as a reviewer for these venues. During his PhD, he was  recipient of multiple scholarships, and  runner-up for the Canon Extreme Imaging Competition in 2015. 
Currently, he is a Research Fellow at UWA since July 2017. Previously, he has also served on the same position at the Australian National University for one year. His current research interests include adversarial machine learning, action recognition and hyperspectral image analysis.

%received BE degree with distinction in Avionics from the College of Aeronautical Engineering, National University of Sciences and Technology (NUST), Pakistan, in 2007 and M.Sc. degree with distinction in Autonomous Systems from Hochschule Bonn-Rhein-Sieg (HBRS), Sankt Augustin, Germany, in 2012.
%He is currently working toward the Ph.D. degree at The University of Western Australia (UWA) under the supervision of Prof. Ajmal Mian.
%He is a recipient of SIRF scholarship at UWA.
%He has served as a Research Assistant at Research Institute for Microwaves and Millimeter-waves Studies, NUST, Pakistan, from 2007 to 2009 and as a Research Associate at the Department of Computer Science at HBRS, Germany in 2012. 
%His current research is focused on hyperspectral image analysis, with emphasis on sparse representation based techniques.  
\end{IEEEbiography}

\begin{IEEEbiography}[{\includegraphics[width=1in,height=1.25in,clip,keepaspectratio]{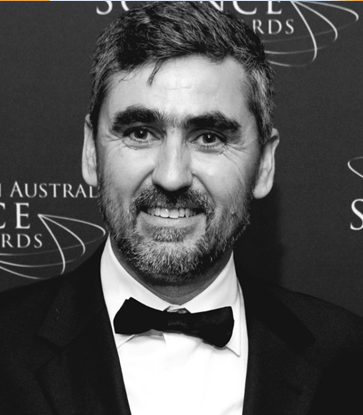}}]{Ajmal Mian} completed his PhD from The University of Western Australia in 2006 with distinction and received the Australasian Distinguished Doctoral Dissertation Award from Computing Research and Education Association of Australasia. He received the prestigious Australian Postdoctoral and Australian Research Fellowships in 2008 and 2011 respectively. He received the UWA Outstanding Young Investigator Award in 2011, the West Australian Early Career Scientist of the Year award in 2012 and the Vice-Chancellors Mid-Career Research Award in 2014. He has secured seven Australian Research Council grants and one National Health and Medical Research Council grant with a total funding of over \$3 Million. He is currently in the School of Computer Science and Software Engineering at The University of Western Australia and is a guest editor of Pattern Recognition, Computer Vision and Image Understanding and Image and Vision Computing journals. His research interests include computer vision, machine learning, 3D shape analysis, hyperspectral image analysis, pattern recognition, and multimodal biometrics.
\end{IEEEbiography}

%\newpage


%\vfill


\end{document}